\documentclass{article}

\usepackage[preprint]{neurips_2024}

\usepackage[utf8]{inputenc} % allow utf-8 input
\usepackage[T1]{fontenc}    % use 8-bit T1 fonts
\usepackage{hyperref}       % hyperlinks
\usepackage{url}            % simple URL typesetting
\usepackage{booktabs}       % professional-quality tables
\usepackage{amsfonts}       % blackboard math symbols
\usepackage{nicefrac}       % compact symbols for 1/2, etc.
\usepackage{microtype}      % microtypography
\usepackage{xcolor}         % colors

\usepackage[small]{caption}
\usepackage{graphicx}
\usepackage{amsmath}
\usepackage{amsthm}
\usepackage{booktabs}
\usepackage{algorithm}
\usepackage{algorithmic}
\usepackage{wrapfig}

\usepackage{algorithm}
\usepackage{algorithmic}
\usepackage{multirow}
\usepackage{colortbl}
\usepackage{makecell}
\usepackage{newfloat}
\usepackage{listings}
\usepackage{booktabs}
\usepackage{amsmath}
\usepackage{mathrsfs} 
\usepackage{subfigure}
\usepackage{multirow}

\title{Tackling Intertwined Data and Device Heterogeneities in Federated Learning with Unlimited Staleness}

\author{%
	Haoming Wang and Wei Gao \\
	University of Pittsburgh\\
	\texttt{hw.wang, weigao@pitt.edu} \\
}

\begin{document}

\maketitle

\begin{abstract}
Federated Learning (FL) can be affected by data and device heterogeneities, caused by clients' different local data distributions and latencies in uploading model updates (i.e., staleness). Traditional schemes consider these heterogeneities as two separate and independent aspects, but this assumption is unrealistic in practical FL scenarios where these heterogeneities are intertwined. In these cases, traditional FL schemes are ineffective, and a better approach is to convert a stale model update into a unstale one. In this paper, we present a new FL framework that ensures the accuracy and computational efficiency of this conversion, hence effectively tackling the intertwined heterogeneities that may cause unlimited staleness in model updates. Our basic idea is to estimate the distributions of clients' local training data from their uploaded stale model updates, and use these estimations to compute unstale client model updates. In this way, our approach does not require any auxiliary dataset nor the clients' local models to be fully trained, and does not incur any additional computation or communication overhead at client devices. We compared our approach with the existing FL strategies on mainstream datasets and models, and showed that our approach can improve the trained model accuracy by up to 25\% and reduce the number of required training epochs by up to 35\%. Source codes can be found at: \url{https://github.com/pittisl/FL-with-intertwined-heterogeneity}.
\end{abstract}

%\vspace{-0.1in}
\section{Introduction}

Federated Learning (FL) \cite{c:6} could be affected by both data and device heterogeneities. \emph{Data heterogeneity} is the heterogeneity of non-i.i.d. data distributions on different clients, which make the aggregated global model biased and reduces model accuracy \cite{c:15,c:1}. \emph{Device heterogeneity} arises from clients' variant latencies in uploading their local model updates to the server, due to their different local resource conditions (e.g., computing power, network link speed, etc). An intuitive solution to device heterogeneity is asynchronous FL, which does not wait for slow clients but updates the global model whenever having received a client update \cite{c:2}. In this case, if a slow client's excessive latency is longer than a training epoch, it will use an outdated global model to compute its model update, which will be \emph{stale} when aggregated at the server and affect model accuracy. To tackle \emph{staleness}, weighted aggregation can be used to apply reduced weights on stale model updates \cite{c:25,c:26}.

Most existing work considers data and device heterogeneities as two separate and independent aspects in FL \cite{c:3}. This assumption, however, is unrealistic in many FL scenarios where these two heterogeneities are \emph{intertwined}: data in certain classes or with particular features may only be available at some slow clients. For example, in FL for hazard rescue \cite{ahmed2020active}, only devices at hazard sites have crucial data about hazards, but they usually have limited connectivity or computing power to timely upload model updates. Similar situations could also happen in FL scenarios where data with high importance to model accuracy is scarce and hard to obtain, such as disease evaluation in smart health, where only few patients have severe symptoms but are very likely to report symptoms with long delays due to their worsening conditions \cite{chen2017pdassist}. 

In these cases, if reduced weights are applied to stale model updates from slow clients, important knowledge in these updates will not be sufficiently learned and hence affects model accuracy. Instead, a better approach is to equally aggregate all model updates and convert a stale model update into a unstale one, but existing techniques for such conversion are limited to a small amount of staleness. For example, first-order compensation can be applied on the gradient delay \cite{zheng2017asynchronous}, by assuming staleness in FL is smaller than one epoch to ignore all the high-order terms in the difference between stale and unstale model updates \cite{c:4}. However, in the aforementioned FL scenarios, it is common to witness excessive or even unlimited staleness, and our experiments in show that the compensation error will quickly increase with staleness.

To efficiently tackle the intertwined data and device heterogeneities with unlimited staleness, in this paper we present a new FL framework that uses gradient inversion at the server to convert stale model updates, by mimicking the local models' gradients produced with their original training data \cite{c:5}. The server inversely computes the gradients from clients' stale model updates to obtain an estimated distribution of clients' training data, such that a model trained with the estimated data distribution will exhibit a similar loss surface as that of using clients' original training data. The server uses such estimated data distributions to retrain the current global model, as estimations of clients' unstale model updates. Compared to other model conversion methods, such as training an extra generative model \cite{c:19} or optimizing input data with constraints \cite{c:18}, our approach has the following advantages:
%\vspace{-0.05in}
\begin{itemize}
	\item Our approach retains the clients' FL procedure to be unchanged, and hence does not incur any additional computation or communication overhead at client devices, which usually have weak capabilities in FL scenarios.
	%\vspace{-0.05in}
	\item Our approach does not require any auxiliary dataset nor the clients' local models to be fully trained, and can hence be widely applied to practical FL scenarios.
	%\vspace{-0.05in}
	\item In our approach, the server will not be able to recover any original samples or labels of clients' local training data. and it cannot produce any recognizable information about clients' local data. Hence, our approach does not impair the clients' data privacy.
\end{itemize}
%\vspace{-0.05in}

We evaluated our proposed technique by comparing with the mainstream FL schemes on multiple datasets and models. Experiment results show that when tackling intertwined data and device heterogeneities with unlimited staleness, our technique can significantly improve the trained model accuracy by up to 25\% and reduce the required amount of training epochs by up to 35\%. Since clients in FL need to compute and upload model updates to the server in every training epoch, such reduction of training epochs largely reduces the computing and communication overhead at clients.

%More technical details about our approach and experiment results can be found in our extended version on ArXiv: \url{https://arxiv.org/abs/2309.13536}. In the rest of this paper, we will refer to different sections of the Technical Appendix, which can also be found in this extended version.

%%\vspace{-0.05in}

\section{Background and Motivation}
In this section, we present preliminary results that demonstrate the ineffectiveness of existing methods in tackling intertwined data and device heterogeneities, hence motivating our proposed approach using gradient inversion.

\begin{wrapfigure}{r}{2.5in}
	\centering
	\vspace{-0.15in}
	\includegraphics[width=2.5in]{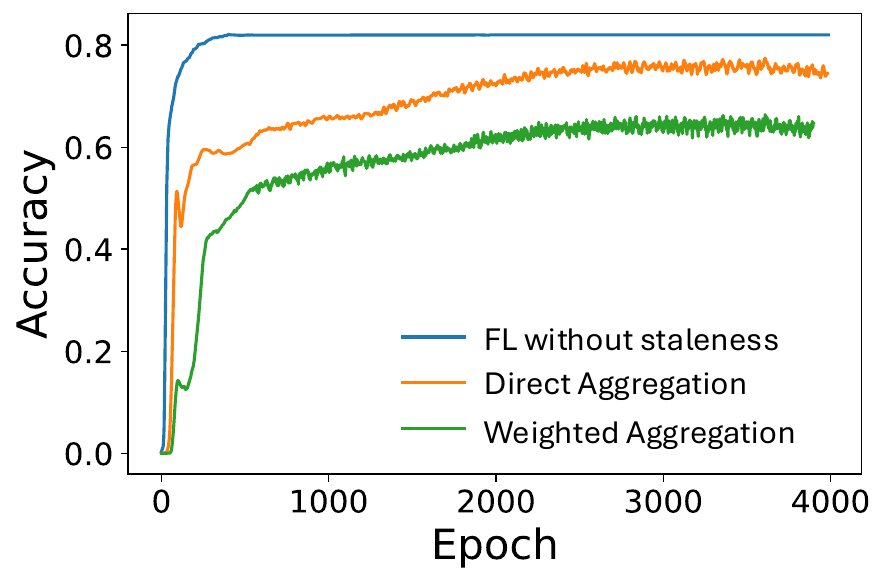} 	
	\vspace{-0.1in}
	\caption{The impact of staleness in FL}
	\vspace{-0.1in}
	\label{fig_weighted_aggregation}
\end{wrapfigure}

%\vspace{-0.05in}
\subsection{Tackling Intertwined Heterogeneities in FL}

Most existing solutions to staleness in AFL are based on weighted aggregation \cite{c:25,c:26,c:27}. For example, [\citenum{c:25}] suggests that a model update's weight exponentially decays with its amount of staleness, and some others use different staleness metrics to decide model updates' weights \cite{c:26}. \cite{c:27} decides these weights based on a feature learning algorithm. These existing solutions, however, are improperly biased towards fast clients, and will hence affect the trained model's accuracy when data and device heterogeneities in FL are intertwined, because they will miss important knowledge in slow clients' model updates. 

To show this, we conducted experiments using a real-world disaster image dataset \cite{mouzannar2018damage}, which contains 6k images of 5 disaster classes (e.g., fires and floods) with different levels of damage severity. In FL of 100 clients, we set \emph{data heterogeneity} as that each client only contain samples in one data class, and set \emph{device heterogeneity} as a staleness of 100 epochs on 15 clients with images of severe damage. When using this dataset to fine-tune a pre-trained ResNet18 model, results in Figure \ref{fig_weighted_aggregation} show that staleness leads to large degradation of model accuracy, and weighted aggregation results in even lower accuracy than direct aggregation, because contributions from images of severe damage on stale clients are reduced by the weights\footnote{In synchronous FL, stale updates will be simply skipped, corresponding to applying zero weights on these updates. Hence, similar performance degradation is also expected for synchronous FL.}.

On the other hand, if we increase the contributions from stale clients by using larger weights, although the model accuracy on these images of severe damage will improve, the larger weights will amplify the impacts of errors contained in stale model updates and hence affect the model's overall accuracy in other data classes. Detailed results can be found in Appendix B.

In practical scenarios such as natural disasters, such large or unlimited staleness is common due to interruptions in communication at disaster sites, and the staleness is too large for server to wait for any slow clients. The large performance degradation of weighted aggregation, then, motivates us to instead convert stale model updates to unstale ones.

%Such large staleness makes it infeasible for the server to wait for slow clients; essentially, since the server has no knowledge about data on these slow clients, it does not know which updates it should wait for. 

%add introduction to first order compensation
% replace 'DC-ASGD'  with 'First-order Compensation' to make it consistent with experiments section
%
The only existing work on such conversion, to our best knowledge, uses the first-order Taylor expansion to compensate for errors in stale model updates \cite{zheng2017asynchronous}. For a stale update $g(w_{t-\tau})$, the estimated unstale update is:
%\vspace{-0.05in}
\begin{equation}
g(w_{t}) \approx g(w_{t-\tau})+\nabla g(w_{t-\tau})(w_{t}-w_{t-\tau}).
%\vspace{-0.05in}
\end{equation}
 Since the Hessian matrix $\nabla g(w_{t-\tau})$ is difficult to compute for neural networks, it is approximated as
%\vspace{-0.05in}
\begin{equation}
\nabla g(w_{t-\tau}) \approx \lambda \cdot g(w_{t-\tau})\odot g(w_{t-\tau})
%\vspace{-0.05in}
\end{equation}
where $\lambda$ is an empirical hyper parameter. However, this method can only applies to small amounts of staleness \cite{zhou2021communication,li2023fedlga,tian2021towards}, in which the high-order terms in the Taylor expansion can be negligible. To verify this, we use the same experiment setting as above and vary the amount of staleness from 0 to 50 epochs. As shown in Table \ref{asgd_error}, the error caused by high-order terms in Taylor expansion, measured in cosine distance and L1-norm difference with the unstale model updates, both significantly increase when staleness increases. These results motivate us to design techniques that ensure accurate conversion with unlimited staleness.

\begin{table}[h]
	\centering
%	\vspace{-0.05in}
	\begin{tabular}{c||ccccc}
		\hline
		Staleness (epoch)  & 5 & 10 & 20 & 50 \\
		\hline 
            \hline  
            Cos-dist error   &0.08 &0.22 & 0.33 & 0.53  \\
            \hline  
		L1-norm error &0.009 &0.018 &0.31& 0.052\\
		\hline
	\end{tabular}
        \vspace{0.05in}
	\caption{Errors caused by high-order terms in Taylor expansion when using \cite{zheng2017asynchronous}}	
	\label{asgd_error}
%	\vspace{-0.1in}
\end{table}

%

%\vspace{-0.1in}
\subsection{Gradient Inversion}

Our proposed approach builds on the existing techniques of gradient inversion \cite{c:9}, which recovers the original training data from the gradient of a trained model. Its basic idea is to minimize the difference between the trained model's gradient and the gradient computed from the recovered data. Denote a batch of training data as $(x,y)$ where $x$ denotes input data and $y$ denotes labels, gradient inversion solves the following optimization problem:
%\vspace{-0.05in}
\begin{equation}
(x'^{*},y'^{*})=\arg \min\nolimits_{(x',y')}\|\frac{\partial L[(x',y');w^{t-1}}{\partial  w^{t-1}}-g^t\|^2_2,
\label{eq:GI}
%\vspace{-0.05in}
\end{equation}
where $(x',y')$ is the recovered data, $w^{t-1}$ is the trained model, $L[\cdot]$ is model's loss function, and $g^t$ is the gradient calculated with training data and $w^{t-1}$. This problem can be solved using gradient descent to iteratively update $(x',y')$.

% change
The quality of recovered data relates to the amount of data samples recovered. Recovering a larger dataset will confuse the learned knowledge across different data samples and reduce the quality of recovered data, and existing methods are limited to recovering a small batch ($<$48) of data samples \cite{c:10,c:11,c:12}. This limitation, however, contradicts with the typical size of clients' datasets in FL, which are usually more than hundreds of samples \cite{wu2023personalized,reddi2020adaptive}. This limitation indicates that we may utilize gradient inversion to estimate clients' training data distributions without revealing individual samples of clients' local data.

\begin{figure*}[ht]
	\centering
	%\vspace{-0.25in}
	%	\hspace{-0.2in}
	\includegraphics[width=\textwidth]{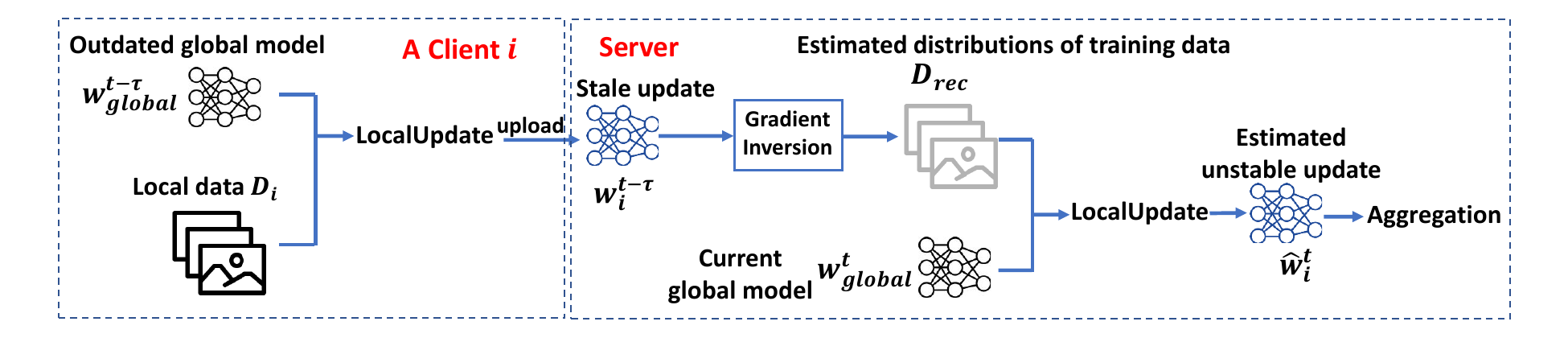} 
	%\vspace{-0.15in}
	\caption{Our proposed method of tackling intertwined data and device heterogeneities in FL}
	\label{fig:flowchart}
	%\vspace{-0.1in}
\end{figure*}

\section{Method}

In this paper, we consider a \emph{semi-asynchronous} FL scenario where some normal clients follow synchronous FL and some slow clients update asynchronously \cite{c:24}. In this case, we measure staleness by the number of epochs that slow clients' updates are delayed. At time $t$\footnote{In the rest of this paper, without loss of generality, we use the notation of time $t$ to indicate the $t$-th epoch in FL training.}, a normal client $i$ provides its model update as 
%\vspace{-0.05in}
\begin{equation}
w^t_i = Local Update (w^t_{global};D_i),
%\vspace{-0.05in}
\end{equation}
where $Local Update[\cdot]$ is client $i$'s local training program, which uses the current global model $w^t_{global}$ and client $i$'s local dataset $D_i$ to produce $w^t_i$. When the client $i$'s model update is delayed, the server will receive a stale model update from $i$ at time $t$ as 
%\vspace{-0.05in}
\begin{equation}
w^{t-\tau}_i = Local  Update (w^{t-\tau}_{global};D_i),
%\vspace{-0.05in}
\end{equation}
where the amount of staleness is indicated by $\tau$ and $w^{t-\tau}_i$ is computed from an outdated global model $w^{t-\tau}_{global}$.

Due to intertwined data and device heterogeneities, we consider that the received $w^{t-\tau}_i$ contains unique knowledge about $D_i$ that is only available from client $i$, and such knowledge should be sufficiently incorporated into the global model. To do so, as shown in Figure \ref{fig:flowchart}, the server uses gradient inversion described in Eq. (\ref{eq:GI}) to recover an intermediate dataset $D_{rec}$ from $w^{t-\tau}_i$. Being different from the existing work of gradient inversion \cite{c:9} that aims to fully recover the client $i$'s training data $D_i$, we only expect $D_{rec}$ to represent the similar data distribution with $D_i$. 

The server then computes an estimation of $w^t_i$ from $w^{t-\tau}_i$, namely $\hat{w}^t_i$, by using $D_{rec}$ to train its current global model $w^t_{global}$, and aggregates $\hat{w}^t_i$ with model updates from other clients to update its global model in the current epoch. %During this procedure, the server only receives the stale model update $w^{t-\tau}_i$ from client $i$, which does not expose $i$'s local data $D_i$ to the server. The client $i$ does not need to perform any extra computations for such estimation of $\hat{w}^t_i$, either.
During this procedure, the server only receives the stale model update $w^{t-\tau}_i$ from client $i$, and we demonstrated that the server's estimation of clients' data distribution will not expose any recognizable information about the clients' local training data, hence avoiding the possible data privacy leakage at clients. At the same time, the computing costs at the client $i$ remains the same as that in vanilla FL, and no any extra computation is needed for such estimation of $\hat{w}^t_i$.

\subsection{Estimating Local Data Distributions from Stale Model Updates}

To compute $D_{rec}$, we first fix the size of $D_{rec}$ and randomly initialize each data sample and label in $D_{rec}$. Then, we iteratively update $D_{rec}$ by minimizing 
%\vspace{-0.05in}
\begin{equation}
Disparity[Local Update (w^{t-\tau}_{global};D_{rec}),w^{t-\tau}_i],
%\vspace{-0.05in}
\label{eq:GI2}
\end{equation}
\begin{figure}[ht]
	\centering
	%	\hspace{-0.15in}
	\vspace{-0.05in}
	\includegraphics[width=0.9\columnwidth]{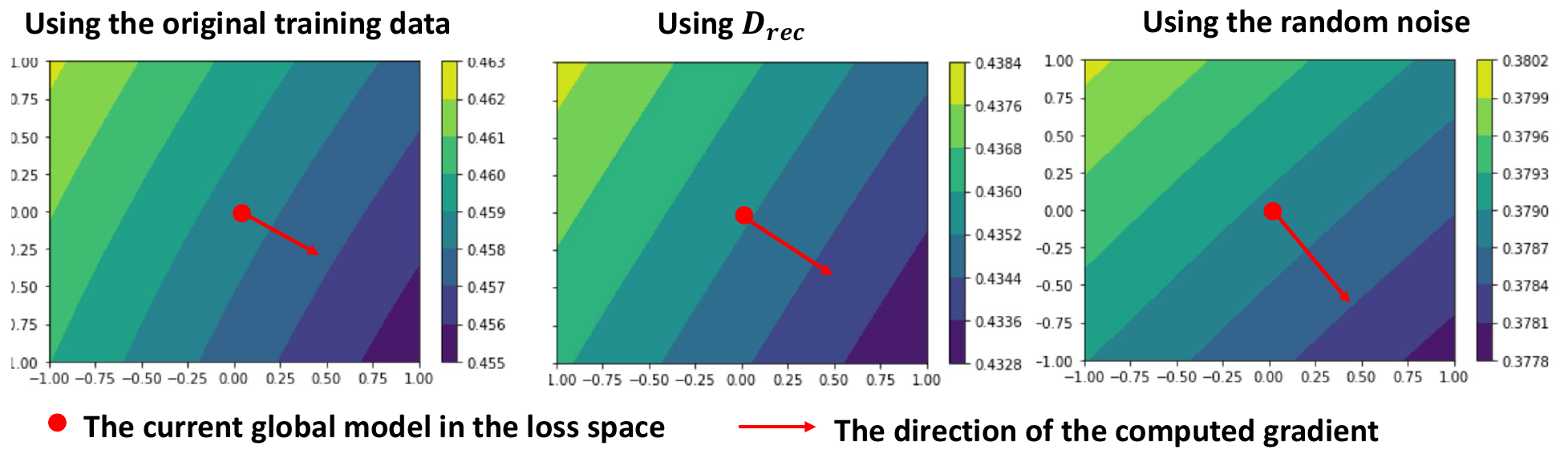} 
	%\vspace{-0.2in}
	\caption{Visualization the loss surface and gradient computed using $D_{rec}$, $D_i$, and random noise data}
	\vspace{-0.1in}
	\label{fig:visualization}
\end{figure}
using gradient descent, where $Disparity[\cdot]$ is a metric to evaluate how much $w^{t-\tau}_i$ changes if being retrained using $D_{rec}$. In FL, a client's model update comprises multiple local training steps instead of a single gradient. Hence, to use gradient inversion in FL, we substitute the single gradient computed from $D_{rec}$ in Eq. (\ref{eq:GI}) with the local training outcome using $D_{rec}$. In this way, since the loss surface in the model's weight space computed using $D_{rec}$ is similar to that using $D_i$, we can expect a similar gradient being computed.

We first visualize it by using MNIST dataset to train LeNet model. Figure \ref{fig:visualization} shows that, the loss surface computed using $D_{rec}$ is similar to that using $D_i$ in the proximity of ($w^{t-\tau}_{global}$), and the computed gradient is very similar, too. 

\begin{wrapfigure}{r}{3in}
	\centering
	\vspace{-0.2in}
	\hspace{-0.15in}
	\subfigure[Cosine distance] { 
		\includegraphics[height=0.17\textwidth]{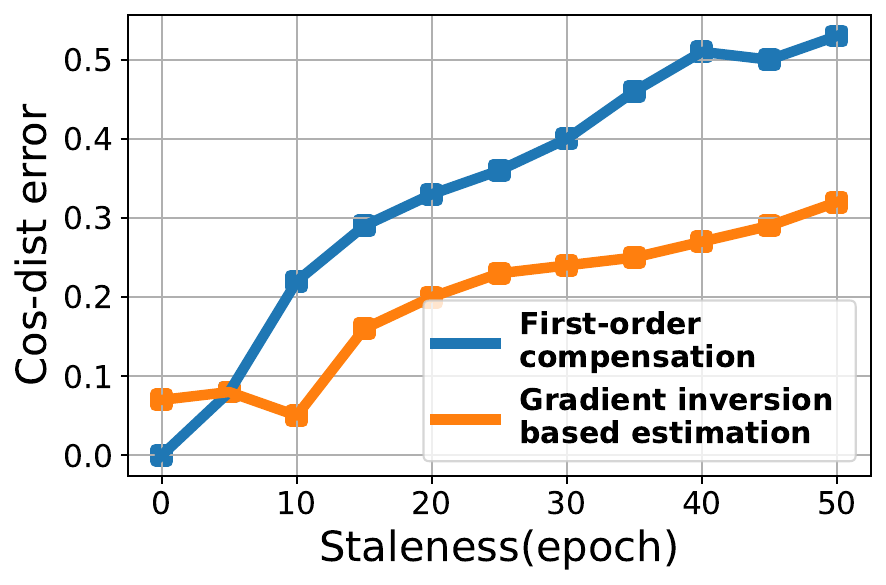}
	}
	\hspace{-0.1in}
	\subfigure[L1-norm difference] { 
		\includegraphics[height=0.17\textwidth]{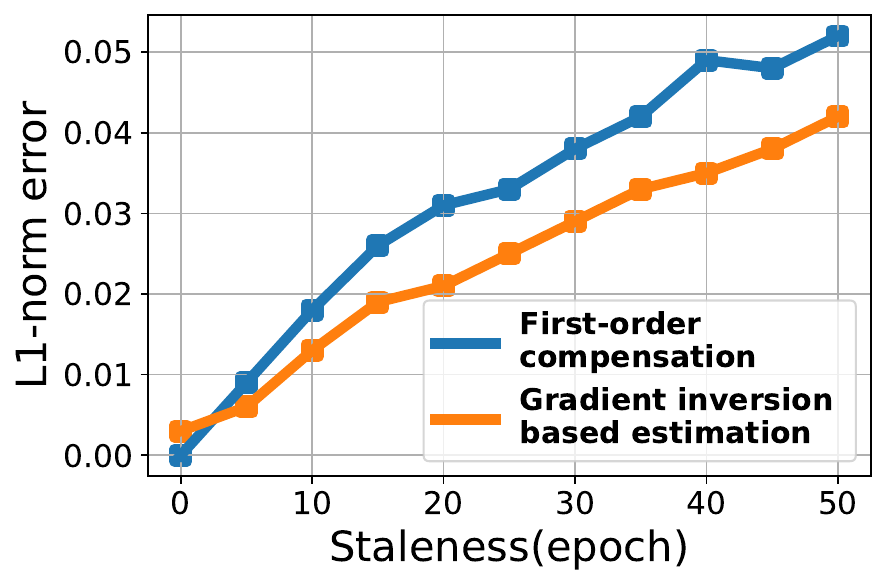}
	}
	\hspace{-0.2in}
%	\vspace{-0.15in}
	\caption{Our method of gradient inversion based estimation has smaller error compared to that of first-order estimation}
%	\vspace{-0.1in}
	\label{error_comparison}
\end{wrapfigure}

To verify the accuracy of using $\hat{w}^t_i$ to estimate $w^t_i$, we compare this estimation with first-order estimation, by computing their discrepancies with the true unstale model update under different amounts of staleness. Results in Figure \ref{error_comparison} show that, compared to First-order Compensation\cite{zheng2017asynchronous}, our estimation based on gradient inversion can reduce the estimation error by up to 50\%, especially when staleness excessively increases to more than 50 epochs.

Another key issue is how to decide the size of $D_{rec}$. Since gradient inversion is equivalent to data resampling in the original training data's distribution, a sufficiently large size of $D_{rec}$ is necessary to ensure unbiased data sampling and sufficient minimization of gradient loss through iterations. On the other hand, when the size of $D_{rec}$ is too large, the computational overhead of each iteration would be unnecessarily too high. More details about how to decide the size of $D_{rec}$ are in Appendix D. Further results about our method's error with various local training programs can also be found in Appendix E.

%%\vspace{-0.05in}
\subsection{Switching back to Vanilla FL in Later Stages of FL}
As shown in Figure \ref{error_comparison}, the estimation made by gradient inversion also contains errors, because the gradient inversion loss can not be reduced to zero. As the FL training progresses and the global model converges, the difference between the previous and current global models will reduce to 0, and hence the difference between stale and unstale model updates will also reduce, eventually to 0. In this case, in the late stage of FL training, the error in our estimated model update ($\hat{w}_i^t$) will exceed that of the original stale model update $w^{t-\tau}_i$.

% compare it with global model update magnitude
\begin{wrapfigure}{r}{3in}
	\centering
	\vspace{-0.2in}
	\hspace{-0.15in}
	\subfigure[Cosine Distance] { 
		\includegraphics[width=0.27\textwidth]{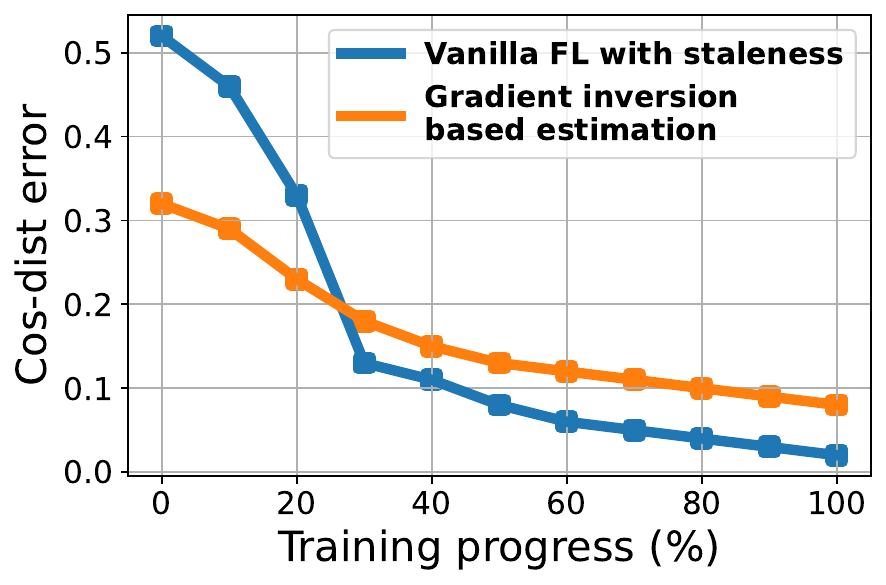}
	}
	\hspace{-0.1in}
	\subfigure[L1-norm difference] { 
		\includegraphics[width=0.27\textwidth]{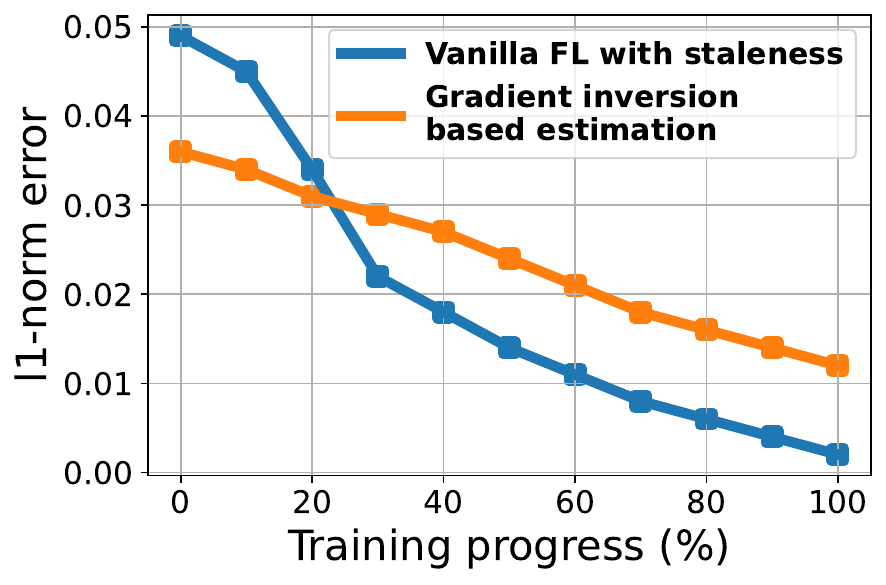}
	}
	\hspace{-0.1in}	%\vspace{-0.1in}
	\vspace{-0.1in}
	\caption{Comparison of model updates' estimation error as the FL training progresses}
	\label{error_over_training}
	\vspace{-0.1in}
\end{wrapfigure}

To verify this, we conducted experiments by training the LeNet model with the MNIST dataset, and evaluated the average values of $E_1(t)=Disparity[\hat{w}^t_i;w^t_i]$ and $E_2(t)=Disparity[w^{t-\tau}_i;w^t_i]$ across different clients, using both cosine distance and L1-norm difference as the metric. Results in Figure \ref{error_over_training} show that at the final stage of FL training, $E_2(t)$ is always larger than $E_1(t)$.

% or switch back to other compensation method

\begin{figure}
	\centering
	\vspace{-0.2in}
	\includegraphics[width=2.2in]{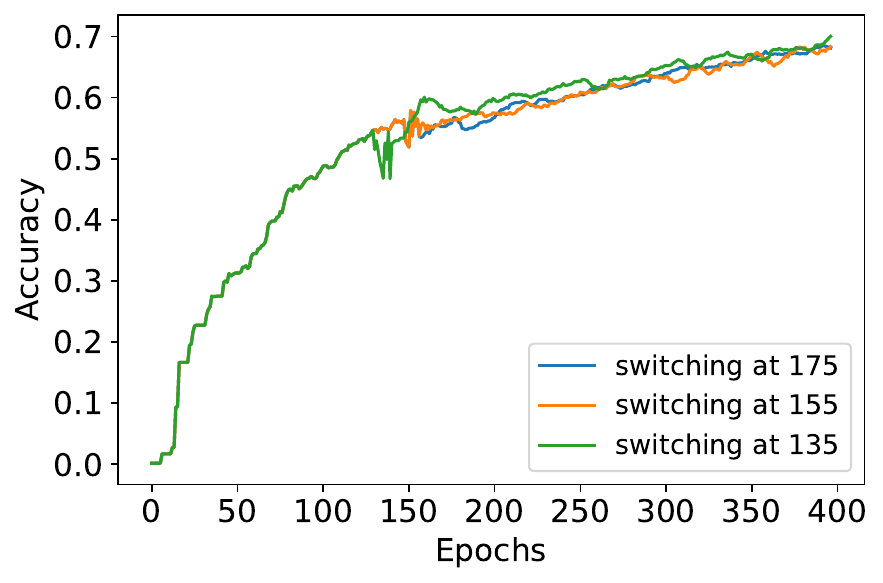}
	\vspace{-0.05in}
	\caption{FL training results with different switching points. $E_1(t)>E_2(t)$ when $t$=155, but different switching points exhibit very similar training performance.}
	\vspace{-0.05in}
	\label{fig:different_switching_points}
\end{figure}

\noindent\textbf{Deciding the switching point.} Hence, in the late stage of FL training, it is necessary to switch back to vanilla FL and directly use stale model updates in aggregation. The difficulty of deciding such switching point is that the true unstale model update ($w^t_i$) is unknown at time $t$. Instead, the server will be likely to receive $w^t_i$ at a later time, namely  $t+\tau'$. Therefore, if we found that $E_1(t)>E_2(t)$ at time $t+\tau'$ when the server receives $w^t_i$ at $t+\tau'$, we can use $t+\tau'$ as the switching point instead of $t$. Doing so will result in a delay in switching, but our experiment results in Table \ref{different_switching_points} and Figure \ref{fig:different_switching_points} with different switching points show that the FL training is insensitive to such delay. 

\begin{table}[h]
	\centering
	\vspace{-0.05in}
	\begin{tabular}{c||ccccc}
		\hline
		Switch point (epoch)  & None & 135 & 155& 175 \\
		\hline 
            \hline  
           Model accuracy  & 59.3\% &68.1\% & 67.4\% & 67.5\%  \\
            \hline  
	\end{tabular}
        \vspace{0.1in}
	\caption{FL training results with different switching points. $E_1(t)>E_2(t)$ when $t$=155, but different switching points exhibit very similar training performance.}	
	\label{different_switching_points}
	\vspace{-0.1in}
\end{table}

In practice, when we make such switch, the model accuracy in training will experience a sudden drop due to the inconsistency of gradients between $\hat{w}^t_i$ and $w^{t-\tau}_i$. To avoid such sudden drop, at time $t+\tau'$, instead of immediately switching to using $\hat{w}^t_i$ in server's model aggregation, we use a weighted average of $\gamma\hat{w}^t_i + (1-\gamma)w^{t-\tau}_i$ in aggregation, so as to ensure smooth switching. $\gamma$ linearly decays from 1 to 0 within a time window, and the length of this window can be flexibly adjusted to accommodate the optimization of model accuracy. Experiment results in Table \ref{alpha_decay} show that, when this length is set to 10\% of training time before reaching the switching point, the model accuracy is maximized.

\begin{table}[h]
	\centering
	\vspace{-0.05in}
	\begin{tabular}{c||ccccc}
		\hline
		Time of decay  & 0\% & 5\% & 10\%& 20\% \\
		\hline 
            \hline  
            Model accuracy  & 67.4\% &69.0\% & 70.2\% & 69.8\%  \\
            \hline  
	\end{tabular}
       \vspace{0.1in}
	\caption{The time needed for $\gamma$ to decay from 1 to 0}	
	\label{alpha_decay}
	\vspace{-0.2in}
\end{table}

\subsection{Computationally Efficient Gradient Inversion}
\label{subsec:reducing_computing_cost}
Our basic design rationale is to retain the clients' FL procedure to be unchanged, and offload all the extra computations incurred by gradient inversion to the server. In this way, we can then focus on reducing the server's computing cost of gradient inversion, which is caused by the large amount of iterations involved, using the following two methods.

First, we reduce complexity of the objective function in gradient inversion by sparsification, which only involve the important gradients with large magnitude into iterations of gradient inversion. Existing work has verified that gradients in mainstream models are highly sparse and only few gradients have large magnitudes \cite{lin2017deep}. Hence, we use a binary mask $Mask[\cdot]$ to selecting elements in $w^{t-\tau}_{i}$ with the top-$K$ magnitudes and only involve these elements to gradient inversion. As shown in Table \ref{sparsification}, by only involving the top 5\% of gradients, we can reduce around 80\% of computation measured as the number of iterations in gradient inversion, with very slight increase in the error of estimating unstale model updates. Besides, we further explored the impact of such error caused by sparsification on the model accuracy, and results are in Appendix F.

\begin{table}[h]
	\centering
	\vspace{-0.05in}
	\begin{tabular}{c||ccccc}
		\hline
		Sparsification rate  & 0\% &90\% & 95\% & 99\% \\
		\hline 
            \hline  
            Reduction of comput. (\%)   &0\% &68\% &80\%  & 93\%  \\
            \hline  
		Estimation error &0.28 &0.29 &0.31 & 0.57\\
		\hline
	\end{tabular}
        \vspace{0.1in}
	\caption{Reduction of computation and error of estimating unstale model updates, with different sparsification rates}	
	\label{sparsification}
	\vspace{-0.1in}
\end{table}

Since in most cases the clients' local data remains fixed, we do not need to start iterations of gradient inversion every time from a random initialization, but could instead optimize $D_{rec}$ from those calculated in the previous training epochs. Our experiments in Table \ref{iteration_reduced} show that, when the clients' local data remains fixed, we can further reduce the amount of iterations in gradient inversion by another 43\%. Even if such client data is only partially fixed (e.g., changed by 20\%), we can still achieve non-negligible reduction of such iterations. 
\begin{table}[H]
	\centering
	\vspace{-0.05in}
	\begin{tabular}{c||ccccc}
		\hline
		Amount of data changed  & 0\% & 5\% & 20\% & 50\% \\
		\hline  
            \hline 
            Computation saved   &43\% &21\% & 12\% &6\%   \\
		
		\hline
	\end{tabular}
\vspace{0.1in}
	\caption{The number of iterations in gradient inversion with different percentages of changes in clients' local data}	
	\label{iteration_reduced}
	\vspace{-0.1in}
\end{table}

Note that, we only apply gradient inversion to stale model updates containing unique knowledge not present in other model updates. Besides, most FL systems \cite{charles2021large} keep the number of clients in a global round constant. Once such number is sufficient (e.g., 10-50 even for FL with thousands of clients), further increasing such number yields little performance gains but increases overhead and causes catastrophic training failure \cite{ro2022scaling}. Hence, the server's overhead of gradient inversion, even in large-scale FL systems, will not largely increase. Such scalability is further discussed in Appendix G.

%\vspace{-0.05in}
\subsection{Protecting Clients' Data Privacy}

Although we used gradient inversion to estimate local data distributions from stale model updates, in most FL settings, it would be difficult or nearly impossible for the server to recover, either the stale clients' local data or the labels, from the knowledge about such distributions, especially when applying the sparsification method described before.

\begin{wrapfigure}{r}{2.8in}
	\centering
	\vspace{-0.1in}
	%\hspace{-0.15in}
	\includegraphics[width=0.5\columnwidth]{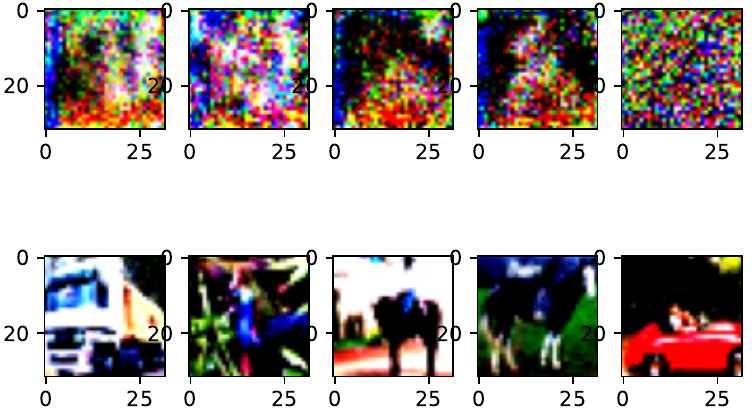} 
%	\vspace{-0.4in}
	\caption{The five best matches between samples in $D_{rec}$ and $D_i$. The top row represents samples in $D_{rec}$, and the bottom row represents samples in $D_i$.}
	\vspace{-0.1in}
	\label{D_rec&D_org}
\end{wrapfigure}

\noindent\textbf{Protecting data samples.} The difficulty of recovering clients' local data samples is proportional to the size of clients' local data and the complexity of local training. In FL, a client's local training data usually contains at least hundreds of samples [\citenum{wang2021field}], and the high diversity among data samples make it difficult to precisely recover any individual sample. To show this, we did experiments with CIFAR-10 dataset and ResNet-18 model, and match each sample in $D_{rec}$ with the most similar sample in $D_i$ based on their LPIPS similarity score [\citenum{c:23}]. As shown in Figure \ref{D_rec&D_org}, these matching data samples are highly dissimilar, and recovered data samples in $D_{rec}$ are mostly meaningless in human perception.

However, even under the easiest scenario where client's dataset only contains one sample and local training is just one-step gradient descent, such recovery will still be unsuccessful. 

\begin{figure}[ht]
	\centering
	\vspace{-0.15in}
	\hspace{-0.3in}
	\subfigure[Original image] { 
		\includegraphics[width=0.19\textwidth]{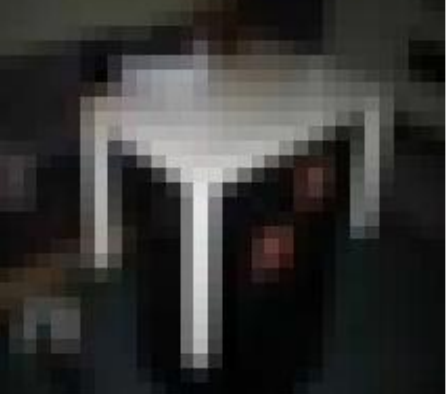}
	}
	%\hspace{-0.05in}
	\subfigure[30\% sparsification] { 
		\includegraphics[width=0.19\textwidth]{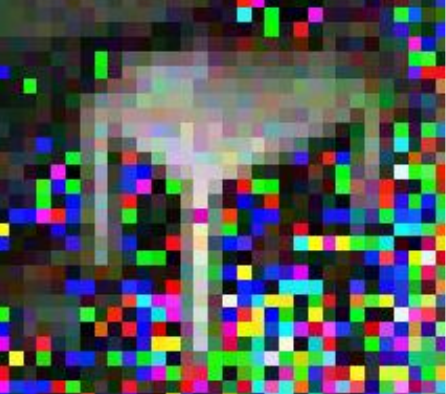}
	}
	%\hspace{-0.05in}	
        \subfigure[75\% sparsification] { 
		\includegraphics[width=0.19\textwidth]{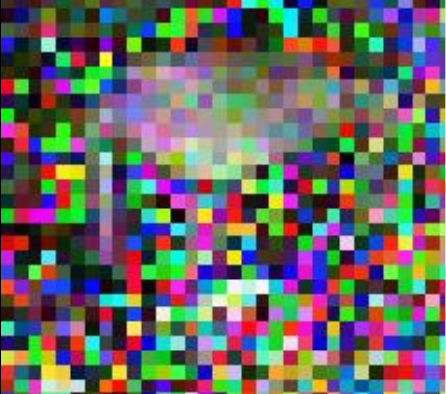}
	}
	%\hspace{-0.05in}
	\subfigure[95\% sparsification] { 
		\includegraphics[width=0.19\textwidth]{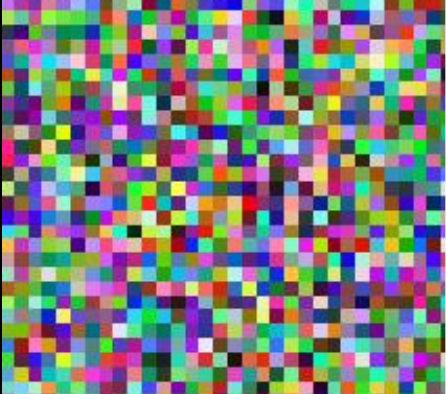}
	}
	\subfigure[Random noise] { 
		\includegraphics[width=0.17\textwidth]{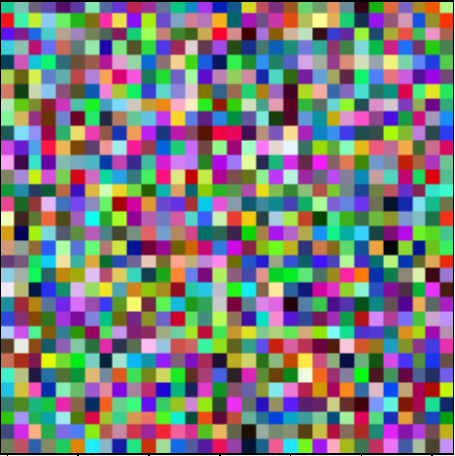}
	}
	\hspace{-0.3in}	
	\vspace{-0.1in}
	\caption{Recovered images under different sparsification rates}
	\label{defense_sparsification}
	\vspace{-0.1in}
\end{figure}

More specifically, although gradient inversion can recover the majority of data samples' pixels as shown in Figure \ref{defense_sparsification}(a) when no sparsification is applied, the quality of such recovery quickly drops when moderate sparsification is applied, as shown in Figure \ref{defense_sparsification}(c) and \ref{defense_sparsification}(d). This is because sparsification effectively reduces the scope of knowledge available for gradient inversion to recover data. Results in Table \ref{Image recovery} with multiple perceptual image quality metrics, including LPIPS \cite{c:23} and FID \cite{heusel2017gans}, further verify that such recovered images cannot be recognized in human eyes. Essentially, when 95\% sparsification rate is applied, the quality of recovered images is similar to that of random noise.
%, and our results in Table \ref{sparsification} verify that the error of estimating unstale model updates with such sparsification rate remains nearly unaffected.
We also assessed the possibility of a neural network classifier (e.g., a ResNet-18 model) to recognize the recovered images. Results in the last row of Table \ref{Image recovery} show that with the 95\% sparsification rate, the classification accuracy is nearly equivalent to random guessing.

Besides, since our method only modifies the FL operations on the server and keeps other FL steps (e.g., the clients' local model updates and client-server communication) unchanged, statistical privacy methods, such as differential privacy, can also be applied to local clients in our approach, just like how it applies to vanilla FL. Each client can independently add Gaussian noise to its local model updates, before sending the updates to the server \cite{geyer2017differentially}. Similarly, it can also apply to our privacy protection method, by adding noise to the gradient after sparsification.

%\begin{table}[ht]
%	\centering
%	  \vspace{-0.05in}
%	\begin{tabular}[ht]{l||cccc}
%		\hline
%			  & 30\% SP & 75\% SP & 95\% SP & Noise \\
%		\hline  
%		\hline
%		MSE $\downarrow$  & 0.014&0.65 & 2.75 & 1.12 \\
%		\hline
%		PSNR $\uparrow$  & 155&77.9 & 41.8 & 47.8 \\
%		\hline
%		LPIPS $\downarrow$  & 0.04&0.13 & 0.56 & 0.50 \\
%		\hline
%		FID $\downarrow$ & 57&102 & 391 & 489 \\
%		\hline
%		ACC$\uparrow$   &87.8& 34.7& 11.2&10.4\\
%		\hline
%	\end{tabular}
%	%\vspace{-0.1in}
%	\caption{The quality of recovered images with different sparsificaition rates (SR) on CIFAR-10 dataset}  
%	\label{Image recovery}
%	\vspace{-0.2in}
%\end{table}

\begin{figure}[ht]
	\centering
	%	  \vspace{-0.15in}
%	{\fontsize{7}{9}\selectfont
		\begin{tabular}{llllllc}
			\toprule
			\textbf{Metric} & \textbf{Model} & \textbf{0\%} & \textbf{30\%} & \textbf{75\%} & \textbf{95\%} & \textbf{Random Noise}\\
			\midrule  
			\multirow{2}{*}{\textbf{MSE} $\downarrow$}& LeNet & 5e-4 & 0.014 & 0.65 & 2.75 & 1.12 \\\cmidrule{2-7}
			& ResNet18 & 0 & 0.011 & 0.87 & 3.16 & 1.12 \\
			\midrule
			\multirow{2}{*}{\textbf{PSNR} $\uparrow$}& LeNet & 261 & 155 & 77.9 & 41.8 & 47.8 \\\cmidrule{2-7}
			& ResNet18 & 323 & 218 & 74.4 & 43.3 & 47.8 \\
			\midrule
			\multirow{2}{*}{\textbf{LPIPS \cite{c:23}} $\downarrow$}& LeNet & 0 & 0.04 & 0.13 & 0.56 & 0.50 \\\cmidrule{2-7}
			& ResNet18 & 0 & 0.01 & 0.18 & 0.59 & 0.50 \\
			\midrule
			\multirow{2}{*}{\textbf{FID \cite{heusel2017gans}} $\downarrow$}& LeNet & 0 & 57 & 102 & 391 & 489 \\\cmidrule{2-7}
			& ResNet18 & 0 & 48 & 114 & 433 & 489 \\
			\midrule
			\multirow{2}{*}{\textbf{Model accuracy (\%)} $\uparrow$}& LeNet & 83.5 & 81.2 & 28.5 & 10.3 & 8.7 \\\cmidrule{2-7}
			& ResNet18 & 89.2 & 87.8 & 34.7 & 11.2 & 10.4 \\
			\bottomrule
	\end{tabular}
%}
	\vspace{0.1in}
	\captionof{table}{Quality of data recovery with different sparsification rates. Different metrics are used to measure the similarity between recovered and original data samples.}  
	\label{Image recovery}
%	\vspace{-0.15in}
\end{figure}

\noindent\textbf{Protecting data labels.} Gradient inversion can be used to recover labels of client's local data \cite{c:9,c:12}.  As shown in Table \ref{Lable recovery}, while such accuracy of label recovery can be as high as 85\% if no protection method is used, applying 95\% sparsification can effectively reduce such accuracy to 66.7\%. Additionally, such accuracy can be further reduced to 46.4\% by adding noise ($var=10^{-3}$) to the gradient, with slight reduction (3\%) of the trained model's accuracy.

\begin{table}[h]
	\centering
	\vspace{-0.05in}
	\begin{tabular}{c||cccc}
		\hline
		Protection method & None & 95\% SP & 95\% SP+noise \\
		\hline  
            \hline 
         Recovery accuracy   &85.5\% &66.7\% & 46.4\%   \\
		
		\hline
	\end{tabular}
\vspace{0.1in}
	\caption{Accuracy of label recovery under different protection methods and sparsification rates (SR)}	
	\label{Lable recovery}
	\vspace{-0.1in}
\end{table}

Gradient inversion should only be applied to stale clients when data and device heterogeneities are intertwined, i.e., the clients' local data is unique and unavailable elsewhere. However, to properly decide such uniqueness, the server will need to know the class labels of client's data, hence impairing the clients' data privacy. Instead, we decide the data uniqueness by comparing the directions of stale clients' model updates with the directions of other model updates from unstale clients, and only consider the stale clients' data as unique if such difference is larger than a given threshold. 
\begin{wrapfigure}{r}{2.2in}
	\centering
	\vspace{-0.1in}
	\includegraphics[width=2.2in]{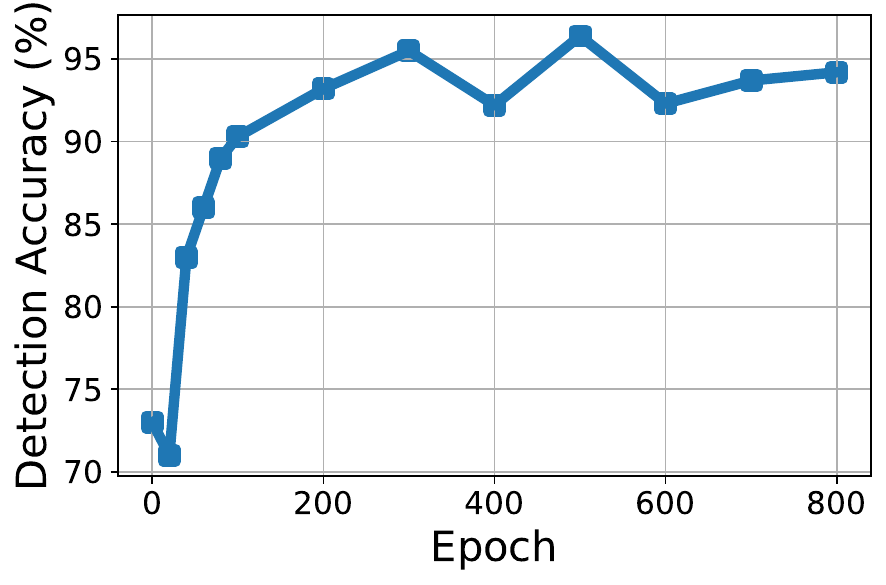}
	\vspace{-0.25in}
	\caption{Accuracy of deciding the uniqueness of clients' local data}
	\vspace{-0.2in}
	\label{fig:detection_acc}
    
\end{wrapfigure}
We quantify such difference between model updates $w_i^t$, $w_j^t$ from client $i$ and $j$ using cosine distance, such that
%\vspace{-0.15in}
\begin{equation}
D_c(w_i^t,w_j^t)=1-w_i^t\cdot w_j^t\biggl{/}\|w_i^t\|\|w_j^t\|,
%\vspace{-0.1in}
\end{equation}
and the threshold is computed as the average of cosine distances between unstale model updates at $t-\tau$:
%\vspace{-0.05in}
\begin{equation}
\frac{1}{\|S^{t-\tau}_{unstale}\|^2} \sum\nolimits_{j,k \in S^{t-\tau}_{unstale}} [D_c(w_j^{t-\tau},w_k^{t-\tau})]
%\vspace{-0.05in}
\end{equation},
where $S^{t-\tau}_{unstale}$ is the set of unstale clients. Since the scale of cosine distance changes during FL training \cite{li2023revisiting}, the average value of cosine distance adds adaptivity to the threshold.

We conducted preliminary experiments to evaluate if the server can accurately detect important model updates from unique client data. In the experiment, we emulate data heterogeneity by assigning each client with data samples from one random class, and results in Table \ref{detection_acc} and Figure \ref{fig:detection_acc} show that the accuracy quickly grows to $>$90\% as training progresses, and the average detection accuracy is 93\%.

\begin{table}[h]
	\centering
	\vspace{-0.05in}
	\begin{tabular}{c||ccccc}
		\hline
		Epoch  & 20 & 100 & 200& 800 \\
		\hline 
            \hline  
            Detection accuracy  &74.6\% &89.2\% & 93.7\% & 94.5\%  \\
            \hline  
	\end{tabular}
        %\vspace{-0.05in}
	\caption{Detection accuracy from stale clients}	
	\label{detection_acc}
	\vspace{-0.2in}
\end{table}

\section{Experiments}
We evaluated our proposed technique in two FL scenarios. In the first scenario, all clients' local datasets are fixed. In the second scenario, we consider a more practical FL setting, where clients' local data is continuously updated and global data distributions are variant over time, due to dynamic changes of environmental contexts. The following baselines that tackle stale model updates in FL are used:

%\vspace{-0.02in}
\begin{itemize}
	\item \textbf{Unweighted aggregation (Unweighted)}: Directly aggregating stale model updates without applying weights.
	\item \textbf{Weighted aggregation (Weighted) \cite{shi2020hysync}}: Applying weights to stale updates in aggregation, and weights are inversely proportional to staleness.
	\item \textbf{First-Order compensation (1st-Order) \cite{zheng2017asynchronous,zhu2022client}}: Compensating errors in stale model updates using first-order Taylor expansion and Hessian approximation.
	\item \textbf{Future global weights prediction (W-Pred) \cite{hakimi2019taming}}: Assuming staleness as pre-known, the future global model is predicted by the first-order method above and used to compensate stale model updates.
	\item \textbf{FL with asynchronous tiers (Asyn-Tiers) \cite{chai2021fedat}}: It clusters clients into asynchronous tiers based on staleness and uses synchronous FL in each tier.
\end{itemize}
%\vspace{-0.02in}
%
FedAvg \cite{c:4} is used in all experiments for aggregating model updates. Hence, Unweighted aggregation is FedAvg with staleness, and Weighted aggregation applies extra weights to model updates in FedAvg\footnote{In FedAvg, updates are also weighted by the number of samples in clients' data, and these two weights are multiplied.}. 1st-Order, W-pred, and our method further modify such weights via compensation, and Asyn-Tiers separately uses FedAvg in each synchronous tier. The usage of FedAvg is independent from our method and other baselines, and can be replaced by other FL frameworks such as FedProx \cite{li2020federated}.

For Weighted aggregation, we set the weights following \cite{shi2020hysync} as $1/(1+e^{a(\tau -b )})$, where $\tau$ is the amount of staleness and we set hyper-parameters $a$=0.25 and $b$=10 based on our experiment settings on staleness. For Asyn-Tiers, we set two asynchronous tiers and when aggregating updates of different tiers, the updates are also weighted by the number of clients in different tiers \cite{chai2021fedat}.

We also evaluated the performance of our technique without staleness, referred as ``\emph{Unstale}'', to assess the disparity between estimated and true values of unstale model updates, as well as the impact of estimation error on FL performance.

%%\vspace{-0.05in}
\subsection{Experiment Setup}
In all experiments, we consider a FL scenario with 100 clients. Each local model update on a client is trained by 5 epochs using the SGD optimizer, with a learning rate of 0.01 and momentum of 0.5.

\begin{figure*}[ht]
	\centering
	\vspace{-0.15in}
	%	\hspace{-0.15in}
	\subfigure[$\alpha$ = 0.1] { 
		\includegraphics[height=0.17\textwidth]{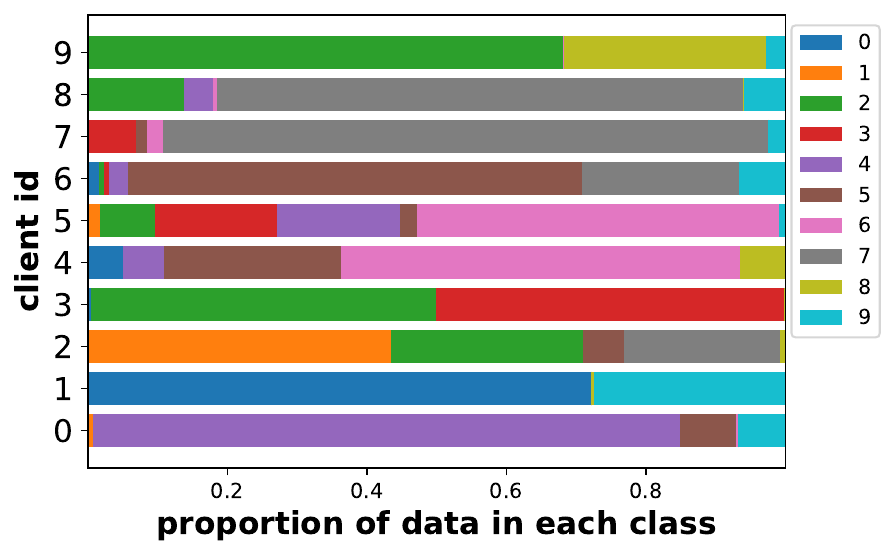}
	}
	\hspace{0.1in}
	\subfigure[$\alpha$ = 1] { 
		\includegraphics[height=0.17\textwidth]{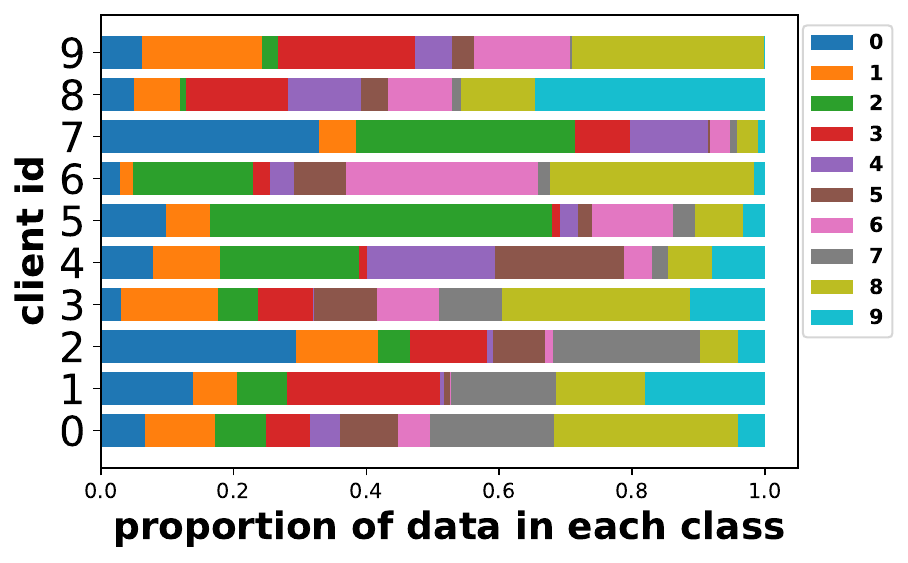}
	}
	\hspace{0.1in}
	\subfigure[$\alpha$ = 100] {\includegraphics[height=0.17\textwidth]{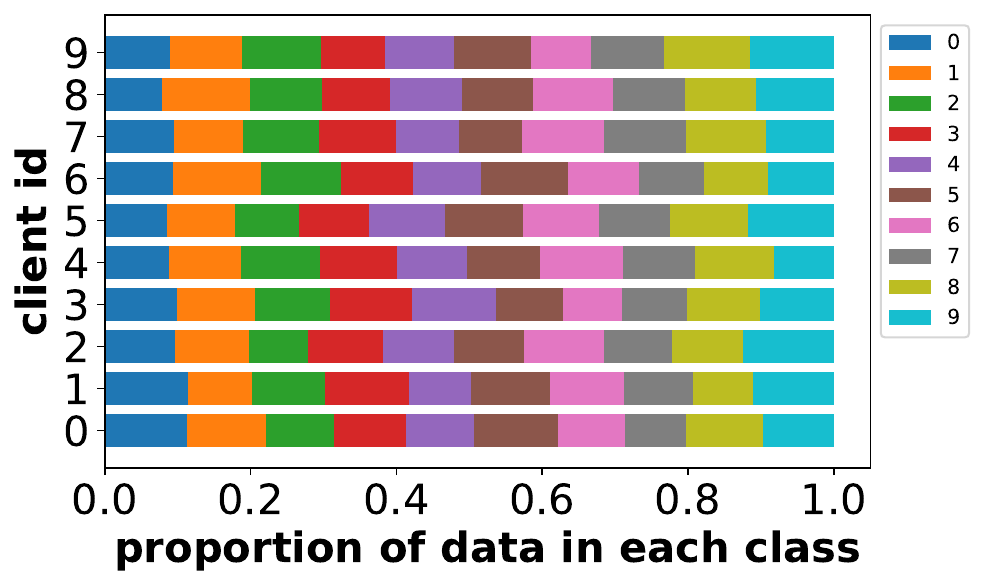}
		
	}
	%	\hspace{-0.1in}
	\vspace{-0.1in}
	\caption{Emulating data heterogeneity using Dirichlet Distribution. Data distributions on 10 clients are shown.}
	\vspace{-0.15in}
	\label{dirichlet}
\end{figure*}

\noindent\textbf{Data heterogeneity}: We use a Dirichlet distribution to sample client datasets with different label distributions \cite{c:29}, and use a tunable parameter ($\alpha$) to adjust the amount of data heterogeneity: as shown in Figure \ref{dirichlet}, the smaller $\alpha$ is, the more biased these label distributions will be and the amount of data heterogeneity is higher. When $\alpha$ is very small, each client only has data samples of few classes.

\noindent\textbf{Device heterogeneity}: To intertwine device heterogeneity with data heterogeneity, we select one data class to be affected by staleness, and apply different amounts of staleness, measured by the number of epochs that clients' model updates are delayed, to the top 10 clients whose local datasets contain the most data samples of the selected data class. The impact of staleness can be further enlarged by applying staleness in the similar way to more data classes.

We evaluate the FL performance by assessing the trained model's accuracy \textbf{in the selected data class} being affected by staleness, and evaluate the FL training time in number of epochs. We expect that our approach can either improve the model accuracy, or achieve the similar accuracy with the baselines but use fewer training epochs.

%\vspace{-0.05in}
\subsection{FL Performance in the Fixed Data Scenario}
In the fixed data scenario, 3 standard datasets and 1 domain-specific dataset are used in evaluations: 
\begin{itemize}
	\item Using MNIST \cite{mnist} and FMNIST \cite{xiao2017fashion} datasets to train a LeNet model, and data class 5 is affected by staleness;
	\item Using CIFAR-10 \cite{cifar10} dataset to train a ResNet-18 model, data class 2 is affected by staleness;
	\item Using a disaster image dataset MDI \cite{mouzannar2018damage} to fine-tune the ResNet-18 model pre-trained with ImageNet.
\end{itemize}

\begin{table}[ht]
	\centering
	\vspace{-0.1in}
	\begin{tabular}{l||l|l|l|l}
		\hline
		Accuracy(\%)  & MNIST & FMNIST & CIFAR10 & MDI \\
		\hline
            \hline
		Unweighted    & 57.4     & 49.2  & 22.8     & 72.3\\
		Weighted    & 39.2     & 30.1  & 12.6 &61.2\\
		1st-Order  & 57.4     & 49.3     & 22.6      &72.3\\
		W-Pred & 57.3     & 48.9      &22.9   &72.2\\
		Asyn-Tiers & 57.6     & 50.3    &25.9   &69.8\\
		\textbf{Ours} & \textbf{61.2}     & \textbf{55.4}   & \textbf{29.4}   & \textbf{75.4}\\
		\hline
	\end{tabular}
\vspace{0.1in}
	\caption{Accuracy of the trained model with different datasets in the fixed data scenario }
	\vspace{-0.1in}
	\label{main}
\end{table}

The trained model's accuracies\footnote{Compared to centralized training, FL models often exhibit lower accuracy, particularly under high data and device heterogeneity, as also reported in existing studies \cite{morafah2024towards}.} using different FL schemes, with the amount of staleness as 40 epochs, are listed in Table \ref{main}. The training progresses of 1st-Order and W-Pred closely resemble that of Unweighted aggregation, suggesting that estimating stale model updates with the Taylor expansion is ineffective under unlimited staleness. Similarly, Weighted aggregation will lead to a biased model with much lower accuracy. In contrast, our gradient inversion based compensation can improve the trained model's accuracy by at least 4\%, compared to the best baseline. Such advantage in model accuracy can be as large as 25\% when compared with Weighted aggregation. Besides image data, our method is also applicable to other data modalities such as text and time-series data. Results and discussions on these modalities with large real-world datasets are in Appendix A.

\begin{figure}[ht]
	\centering
	\vspace{-0.1in}
	\hspace{-0.15in}
	\subfigure[MNIST, LeNet] { 
		\includegraphics[width=0.4\textwidth]{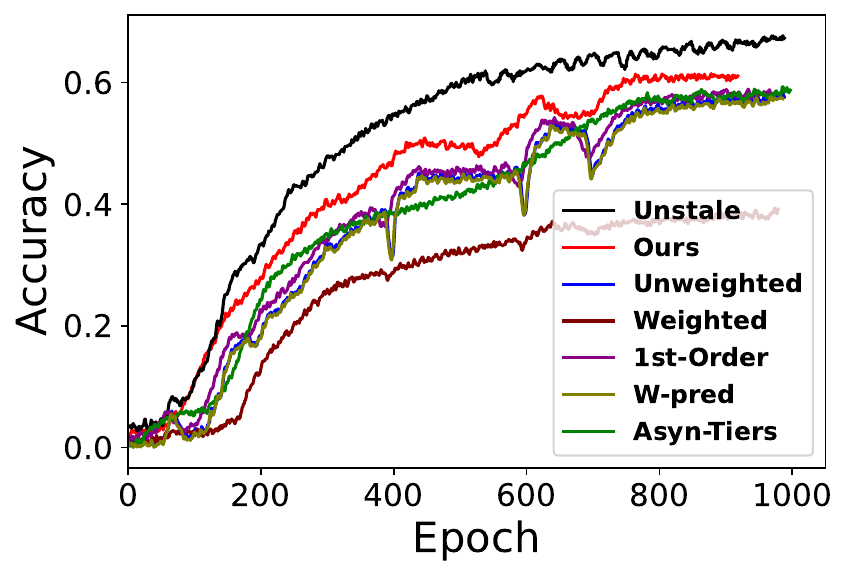}
	}
	\hspace{-0.1in}
	\subfigure[CIFAR-10, ResNet18] { 
		\includegraphics[width=0.4\textwidth]{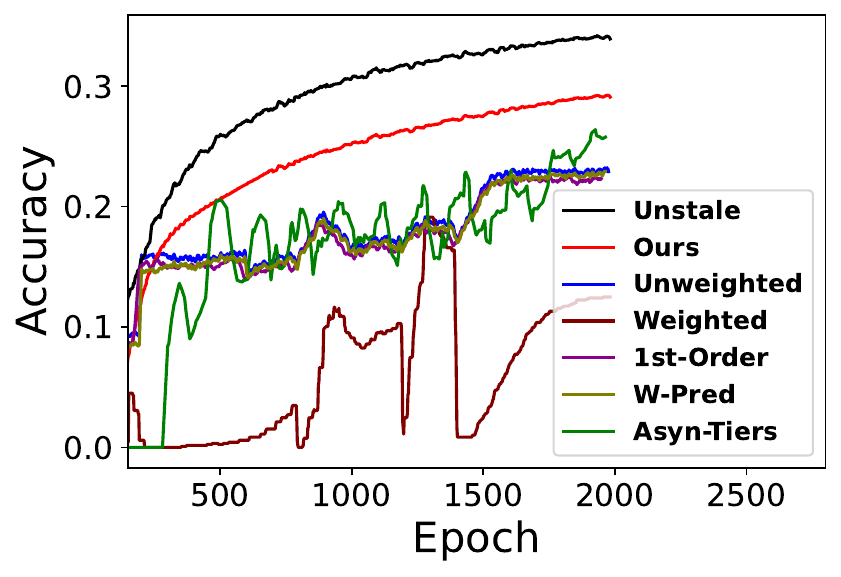}
	}
	\hspace{-0.1in}	\vspace{-0.1in}
	\caption{The FL training procedure}
	\label{fixed_curve}
	\vspace{-0.1in}
\end{figure}

Figure \ref{fixed_curve} further show the FL training procedure over different epochs, and demonstrated that our method can also improve the progress and stability of training while also achieve higher model accuracy during different stages of FL training. Furthermore, we conducted experiments with different amounts of data and device heterogeneity. Results in Tables \ref{variation_alpha_fixed} and \ref{variation_staleness_fixed} show that\footnote{Training times in Tables 10-13 represent the relative training time required to reach convergence of the global model, assuming our method's training time is 100\%.}, compared with the baselines, our method can generally achieve higher model accuracy or reach the same accuracy with fewer training epochs, especially when the amount of staleness is large or the amount of data heterogeneity is high. We also use other large-scale real-world dataset to conduct experiments and results are in Appendix C.

%\iffalse
%\begin{figure}[ht]
%	\centering
%	%\vspace{-0.1in}
%	\hspace{-0.15in}
%	\subfigure[Model accuracy] { 
%		\includegraphics[width=0.23\textwidth]{image/variation_alpha_acc.pdf}
%	}
%	\hspace{-0.1in}
%	\subfigure[Reduction of training epochs] { 
%		\includegraphics[width=0.23\textwidth]{image/variation_alpha_time.pdf}
%	}
%	\hspace{-0.1in}	%\vspace{-0.1in}
%	\caption{Model accuracy and percentage of training epochs being saved, with \emph{different amounts of
%		data heterogeneity} controlled by $\alpha$ in the Dirichlet distribution}
%	\label{variation_alpha_fixed}
%	%\vspace{-0.1in}
%\end{figure}
%
%
%\begin{figure}[ht]
%	\centering
%	%\vspace{-0.1in}
%	\hspace{-0.15in}
%	\subfigure[Model accuracy] { 
%		\includegraphics[width=0.23\textwidth]{image//variation.pdf}
%	}
%	\hspace{-0.1in}
%	\subfigure[Reduction of training epochs] { 
%		\includegraphics[width=0.23\textwidth]{image/variation_staleness_time_fixed.pdf}
%	}
%	\hspace{-0.1in}	%\vspace{-0.1in}
%	\caption{The trained model accuracy and percentage of training epochs being saved, with \textbf{different amounts of staleness} measured in the number of delayed epochs}
%	\label{variation_staleness_fixed}
%	%\vspace{-0.1in}
%\end{figure}
%\fi

\begin{table}[H]
        \small
	\centering
	\vspace{-0.05in}
	\begin{tabular}{c||c|c||c|c||c|c}
		\hline
		$\alpha$ & \multicolumn{2}{c||}{1} &\multicolumn{2}{c||}{0.1}&\multicolumn{2}{c}{0.01} \\
		\hline 
            \hline  
            & Acc & Time & Acc & Time& Acc & Time  \\
            \hline  
		  Unweighted  &82.3 &100 & 57.4&128 & 51.1&132\\
            Weighted    &82.4  & 102&39.2 &171 & 31.1&179 \\
            1st-Order   & 82.5& 100&57.3& 129&51.5 &131\\
            W-Pred      & 82.8& 100& 57.6& 126&50.9 &131\\
            Asyn-tiers  &82.3 & \textbf{97}& 57.6& 126&52.7 &135\\
            \textbf{Ours}        & \textbf{82.3}& 100 & \textbf{61.2} & \textbf{100} & \textbf{58.3} &	\textbf{100}\\
            \hline
	\end{tabular}
       \vspace{0.1in}
	\caption{Model accuracy and percentage of training epochs being saved, with \emph{different amounts of
		data heterogeneity} controlled by $\alpha$ in the Dirichlet distribution. The MNIST dataset and LeNet model are used.}	
	\label{variation_alpha_fixed}
	\vspace{-0.1in}
\end{table}

\begin{table}[H]
        \small
	\centering
	\vspace{-0.05in}
	\begin{tabular}{c||c|c||c|c||c|c}
		\hline
		Staleness  & \multicolumn{2}{c||}{10} &\multicolumn{2}{c||}{40}&\multicolumn{2}{c}{100} \\
		\hline 
            \hline  
            & Acc & Time & Acc & Time& Acc & Time  \\
            \hline  
		  Unweighted &72.6 &104&57.4 &128 &41.5 &142\\
            Weighted   &69.4 &115 &39.2 &171 &30.5 & 179\\
            1st-Order  &72.6 & 104&57.3 & 129& 41.8&141\\
            W-Pred     & 72.6& 104& 57.6& 126&41.7 &142\\
            Asyn-tiers & 72.7& 103&57.6 & 126& 38.3&138\\
            \textbf{Ours} & \textbf{73.3} & \textbf{100} & \textbf{61.2} & \textbf{100} & \textbf{47.2} & \textbf{100}\\
		\hline
	\end{tabular}
       \vspace{0.1in}
	\caption{Model accuracy and percentage of training epochs being saved, with \emph{different amounts of staleness} measured in the number of delayed epochs.}	
	\label{variation_staleness_fixed}
	\vspace{-0.1in}
\end{table}

%\vspace{-0.1in}
\subsection{FL Performance in the Variant Data Scenario}

\vspace{-0.05in}
\begin{wrapfigure}{r}{2.3in}
	\centering
	\vspace{-0.3in}
	\includegraphics[width=0.4\columnwidth]{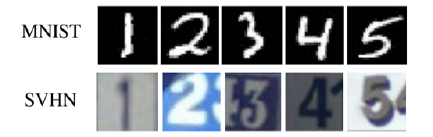} 
	\vspace{-0.05in}
	\caption{Datasets for digit recognition: MNIST and SVHN}
	\vspace{-0.1in}
	\label{mnist_svhn}
\end{wrapfigure}

To continuously vary the global data distributions, we use two public datasets, namely MNIST and SVHN \cite{svhn}, which are for the same learning task but with different feature representations as shown in Figure \ref{mnist_svhn}. Each client's local dataset is initialized as the MNIST dataset in the same way as in the fixed data scenario. Afterwards, during training, each client continuously replaces random data samples in its local dataset with new data samples in the SVHN dataset.

\begin{figure}[H]
	\centering
	\vspace{-0.1in}
	\includegraphics[width=0.4\textwidth]{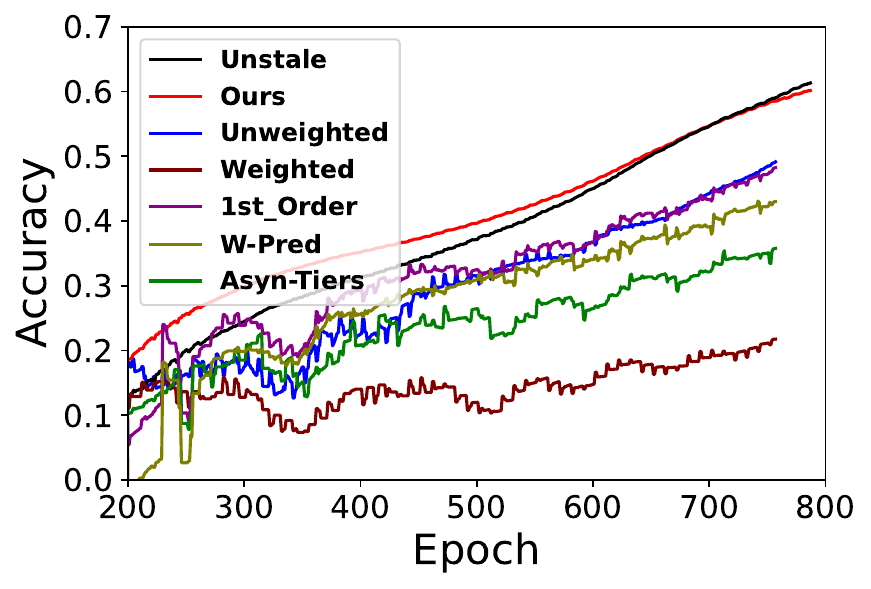} 
	\vspace{-0.1in}
	\caption{Model accuracy with variant data distributions in clients' local datasets}
	\vspace{-0.2in}
	\label{streaming_dataset1}
\end{figure}

Experiment results in Figure \ref{streaming_dataset1} show that in such variant data scenario, since clients' local data distributions continuously change, the FL training will never converge. Hence, the model accuracy achieved by the existing FL schemes exhibited significant fluctuations over time and stayed low ($<$40\%). In comparison, our technique can better depict the variant data patterns and hence achieve much higher model accuracy, which is comparable to FL without staleness and 20\% higher than those in existing FL schemes.

\begin{table}[H]
        \small
	\centering
	\vspace{-0.05in}
	\begin{tabular}{c||c|c||c|c||c|c}
		\hline
		Staleness  & \multicolumn{2}{c||}{10} &\multicolumn{2}{c||}{40}&\multicolumn{2}{c}{100} \\
		\hline 
            \hline  
            & Acc & Time & Acc & Time& Acc & Time  \\
            \hline  
		  Unweighted  &60.6  &99 & 53.2&117 & 39.1&131\\
            Weighted    &59.8  & 109&38.9 &153 & 21.8&166 \\
            1st-Order   & 60.6& 100&53.6& 117&40.0 &133\\
            W-Pred      & 60.4& 100& 53.3& 117&39.1 &131\\
            Asyn-tiers  &58.2 & 103& 46.9& 118&35.7 &137\\
            \textbf{Ours}        & \textbf{63.3} & \textbf{100} & \textbf{62.5} & \textbf{100} & \textbf{61.0} &	\textbf{100}\\
            \hline
	\end{tabular}
       \vspace{0.1in}
	\caption{Model accuracy and the number of training epochs reduced, with \emph{different amounts of
		staleness} measured in the number of delayed epochs}	
	\label{variation_staleness_streaming}
	\vspace{-0.1in}
\end{table}

We also conducted experiments with different amounts of staleness and different rates of data distributions' variations. We apply different variation rates of clients' local data distributions by replacing different amounts of such random data samples in the clients' local datasets in each epoch. To prevent the training from stopping too early when the variation rate is high, we repeatedly varied the data when the variation rate exceeded 1 sample per epoch. Results in Tables \ref{variation_staleness_streaming} and \ref{variation_staleness_rate} showed that our method outperformed the baselines with different amounts of staleness. Weighted aggregation performs the worst since it leads the model to bias toward other unstable clients and other baselines show similar performance since they cannot compensate such large staleness.

\begin{table}[H]
        \small
	\centering
	\vspace{-0.05in}
	\begin{tabular}{c||c|c||c|c||c|c}
		\hline
		Rate & \multicolumn{2}{c||}{0.5} &\multicolumn{2}{c||}{1}&\multicolumn{2}{c}{2} \\
		\hline 
            \hline  
            & Acc & Time & Acc & Time& Acc & Time  \\
            \hline  
		  Unweighted  &73.1  &100 & 39.1&131 & 44.1&127\\
            Weighted    &58.2& 102&21.8 &166 & 25.2&163 \\
            1st-Order   & 73.2& 100&40.0& 133&43.9 &127\\
            W-Pred      & 73.1& 101& 39.0& 131&39.5 &127\\
            Asyn-tiers  &68.3 & 98& 35.7& 137&39.1 &130\\
            \textbf{Ours}        & \textbf{70.3} & \textbf{100} & \textbf{60.1} & \textbf{100} & \textbf{63.3} & \textbf{100}\\
            \hline
	\end{tabular}
      \vspace{0.1in}
	\caption{Model accuracy and the number of training epochs reduced, with \emph{different rates of data distributions' variations} measured by number of samples changed per epoch}	
	\label{variation_staleness_rate}
	\vspace{-0.1in}
\end{table}

\section{Related Work}

Most existing solutions to staleness in FL are based on weighted aggregation \cite{c:25,c:26,c:27}. These existing solutions are biased towards fast clients, and will affect the trained model's accuracy when data and device heterogeneities in FL are intertwined. Other researchers suggest to use semi-asynchronous FL, where the server aggregates client model updates at a lower frequency \cite{c:7} or clusters clients into different asynchronous ``tiers'' according to their update rates \cite{c:24}. However, doing so cannot completely eliminate the impact of intertwined data and device heterogeneities, because the server's aggregation still involves stale model updates. 

Instead, we can transfer knowledge from stale model updates to the global model, by training a generative model and compelling its generated data to exhibit high predictive values on the original model updates \cite{c:20,lopes2017data,zhu2021data}. Another approach is to optimize randomly initialized input data until it has good performance on the original model \cite{c:21}. However, the quality and accuracy of knowledge transfer in these methods remains low, and we provided more detailed experiment results in Appendix C to demonstrate such low quality. Other efforts enhance the quality of knowledge transfer by incorporating natural image priors \cite{c:22} or using another public dataset to introduce general knowledge \cite{c:19}, but require extra datasets. Moreover, all these methods require that the clients' local models to be fully trained, which is usually infeasible in FL.

%%\vspace{-0.15in}

%\vspace{-0.1in}
\section{Conclusion}
In this paper, we present a new FL framework to tackle intertwined data and device heterogeneities in FL, by using gradient inversion to estimate clients' unstale model updates. Experiments show that our technique largely improves model accuracy and reduces the amount of training epochs needed.

\section*{Acknowledgments}
We thank the anonymous reviewers for their comments and feedback. This work was supported in part by National Science Foundation (NSF) under grant number IIS-2205360, CCF-2217003, CCF-2215042, and National Institutes of Health (NIH) under grant number R01HL170368.

\setcitestyle{numbers}
\bibliographystyle{abbrvnat}
\bibliography{aaai25}

\begin{thebibliography}{59}
\providecommand{\natexlab}[1]{#1}
\providecommand{\url}[1]{\texttt{#1}}
\expandafter\ifx\csname urlstyle\endcsname\relax
  \providecommand{\doi}[1]{doi: #1}\else
  \providecommand{\doi}{doi: \begingroup \urlstyle{rm}\Url}\fi

\bibitem[Ahmed et~al.(2020)Ahmed, Ahmad, Said, Qolomany, Qadir, and
  Al-Fuqaha]{ahmed2020active}
L.~Ahmed, K.~Ahmad, N.~Said, B.~Qolomany, J.~Qadir, and A.~Al-Fuqaha.
\newblock Active learning based federated learning for waste and natural
  disaster image classification.
\newblock \emph{IEEE Access}, 8:\penalty0 208518--208531, 2020.

\bibitem[Chai(2021)]{c:24}
e.~a. Chai, Zheng.
\newblock {FedAT: A high-performance and communication-efficient federated
  learning system with asynchronous tiers.}
\newblock In \emph{Proceedings of the International Conference for High
  Performance Computing, Networking, Storage and Analysis}, 2021.

\bibitem[Chai et~al.(2021)Chai, Chen, Anwar, Zhao, Cheng, and
  Rangwala]{chai2021fedat}
Z.~Chai, Y.~Chen, A.~Anwar, L.~Zhao, Y.~Cheng, and H.~Rangwala.
\newblock Fedat: A high-performance and communication-efficient federated
  learning system with asynchronous tiers.
\newblock In \emph{Proceedings of the International Conference for High
  Performance Computing, Networking, Storage and Analysis}, pages 1--16, 2021.

\bibitem[Charles et~al.(2021)Charles, Garrett, Huo, Shmulyian, and
  Smith]{charles2021large}
Z.~Charles, Z.~Garrett, Z.~Huo, S.~Shmulyian, and V.~Smith.
\newblock On large-cohort training for federated learning.
\newblock \emph{Advances in neural information processing systems},
  34:\penalty0 20461--20475, 2021.

\bibitem[Charpiat(2019)]{c:28}
e.~a. Charpiat, Guillaume.
\newblock {Input similarity from the neural network perspective.}
\newblock In \emph{Advances in Neural Information Processing Systems 32}, 2019.

\bibitem[Chen(2020)]{c:27}
e.~a. Chen, Yujing.
\newblock {Asynchronous online federated learning for edge devices with non-iid
  data.}
\newblock In \emph{2020 IEEE International Conference on Big Data (Big Data)},
  2020.

\bibitem[Chen and Jin.(2019)]{c:25}
X.~S. Chen, Yang and Y.~Jin.
\newblock {Communication-efficient federated deep learning with layerwise
  asynchronous model update and temporally weighted aggregation.}
\newblock In \emph{IEEE transactions on neural networks and learning systems},
  2019.

\bibitem[Chen et~al.(2017)Chen, Yang, Chen, Miao, and Yu]{chen2017pdassist}
Y.~Chen, X.~Yang, B.~Chen, C.~Miao, and H.~Yu.
\newblock Pdassist: Objective and quantified symptom assessment of parkinson's
  disease via smartphone.
\newblock In \emph{2017 IEEE International Conference on Bioinformatics and
  Biomedicine (BIBM)}, pages 939--945. IEEE, 2017.

\bibitem[Dimitrov et~al.(2022)Dimitrov, Balunovi{\'c}, Jovanovi{\'c}, and
  Vechev]{dimitrov2022lamp}
D.~I. Dimitrov, M.~Balunovi{\'c}, N.~Jovanovi{\'c}, and M.~Vechev.
\newblock Lamp: Extracting text from gradients with language model priors.
\newblock \emph{arXiv e-prints}, pages arXiv--2202, 2022.

\bibitem[Geiping(2020)]{c:11}
e.~a. Geiping, Jonas.
\newblock {Inverting gradients-how easy is it to break privacy in federated
  learning?}
\newblock In \emph{Advances in Neural Information Processing Systems 33}, 2020.

\bibitem[Geyer et~al.(2017)Geyer, Klein, and Nabi]{geyer2017differentially}
R.~C. Geyer, T.~Klein, and M.~Nabi.
\newblock Differentially private federated learning: A client level
  perspective.
\newblock \emph{arXiv preprint arXiv:1712.07557}, 2017.

\bibitem[Gupta et~al.(2022)Gupta, Huang, Zhong, Gao, Li, and
  Chen]{gupta2022recovering}
S.~Gupta, Y.~Huang, Z.~Zhong, T.~Gao, K.~Li, and D.~Chen.
\newblock Recovering private text in federated learning of language models.
\newblock \emph{Advances in neural information processing systems},
  35:\penalty0 8130--8143, 2022.

\bibitem[Hakimi et~al.(2019)Hakimi, Barkai, Gabel, and
  Schuster]{hakimi2019taming}
I.~Hakimi, S.~Barkai, M.~Gabel, and A.~Schuster.
\newblock Taming momentum in a distributed asynchronous environment.
\newblock \emph{arXiv preprint arXiv:1907.11612}, 2019.

\bibitem[Heusel et~al.(2017)Heusel, Ramsauer, Unterthiner, Nessler, and
  Hochreiter]{heusel2017gans}
M.~Heusel, H.~Ramsauer, T.~Unterthiner, B.~Nessler, and S.~Hochreiter.
\newblock Gans trained by a two time-scale update rule converge to a local nash
  equilibrium.
\newblock \emph{Advances in neural information processing systems}, 30, 2017.

\bibitem[Hsu and Brown.(2019)]{c:29}
H.~Q. Hsu, Tzu-Ming~Harry and M.~Brown.
\newblock {Measuring the effects of non-identical data distribution for
  federated visual classification.}
\newblock In \emph{arXiv preprint}, 2019.

\bibitem[Karimireddy(2020)]{c:17}
e.~a. Karimireddy, Sai~Praneeth.
\newblock {Scaffold: Stochastic controlled averaging for federated learning.}
\newblock In \emph{International conference on machine learning. PMLR}, 2020.

\bibitem[Konečný(2016)]{c:15}
e.~a. Konečný, Jakub.
\newblock {Federated optimization: Distributed machine learning for on-device
  intelligence.}
\newblock In \emph{arXiv preprint arXiv:1610.02527}, 2016.

\bibitem[Krizhevsky(2009)]{cifar10}
a.~G.~H. Krizhevsky, Alex.
\newblock \emph{{Learning multiple layers of features from tiny images.}}
\newblock 2009.

\bibitem[LeCun and Burges.(2010)]{mnist}
C.~C. LeCun, Yann and C.~Burges.
\newblock Mnist handwritten digit database.
\newblock \url{http://yann.lecun.com/exdb/mnist}, 2010.

\bibitem[Li et~al.(2020)Li, Sahu, Zaheer, Sanjabi, Talwalkar, and
  Smith]{li2020federated}
T.~Li, A.~K. Sahu, M.~Zaheer, M.~Sanjabi, A.~Talwalkar, and V.~Smith.
\newblock Federated optimization in heterogeneous networks.
\newblock \emph{Proceedings of Machine learning and systems}, 2:\penalty0
  429--450, 2020.

\bibitem[Li et~al.(2023{\natexlab{a}})Li, Qu, Tang, and Lu]{li2023fedlga}
X.~Li, Z.~Qu, B.~Tang, and Z.~Lu.
\newblock Fedlga: Toward system-heterogeneity of federated learning via local
  gradient approximation.
\newblock \emph{IEEE Transactions on Cybernetics}, 54\penalty0 (1):\penalty0
  401--414, 2023{\natexlab{a}}.

\bibitem[Li et~al.(2023{\natexlab{b}})Li, Lin, Shang, and Wu]{li2023revisiting}
Z.~Li, T.~Lin, X.~Shang, and C.~Wu.
\newblock Revisiting weighted aggregation in federated learning with neural
  networks.
\newblock \emph{arXiv preprint arXiv:2302.10911}, 2023{\natexlab{b}}.

\bibitem[Lin et~al.(2017)Lin, Han, Mao, Wang, and Dally]{lin2017deep}
Y.~Lin, S.~Han, H.~Mao, Y.~Wang, and W.~J. Dally.
\newblock Deep gradient compression: Reducing the communication bandwidth for
  distributed training.
\newblock \emph{arXiv preprint arXiv:1712.01887}, 2017.

\bibitem[Lopes et~al.(2017)Lopes, Fenu, and Starner]{lopes2017data}
R.~G. Lopes, S.~Fenu, and T.~Starner.
\newblock Data-free knowledge distillation for deep neural networks.
\newblock \emph{arXiv preprint arXiv:1710.07535}, 2017.

\bibitem[Luo(2020)]{c:22}
e.~a. Luo, Liangchen.
\newblock {Large-scale generative data-free distillation.}
\newblock In \emph{arXiv preprint arXiv:2012.05578}, 2020.

\bibitem[McMahan(2016)]{c:6}
e.~a. McMahan, Brendan.
\newblock {Communication-efficient learning of deep networks from decentralized
  data.}
\newblock In \emph{arXiv preprint}, 2016.

\bibitem[Morafah et~al.(2024)Morafah, Kungurtsev, Chang, Chen, and
  Lin]{morafah2024towards}
M.~Morafah, V.~Kungurtsev, H.~Chang, C.~Chen, and B.~Lin.
\newblock Towards diverse device heterogeneous federated learning via task
  arithmetic knowledge integration.
\newblock \emph{arXiv preprint arXiv:2409.18461}, 2024.

\bibitem[Mouzannar et~al.(2018)Mouzannar, Rizk, and Awad]{mouzannar2018damage}
H.~Mouzannar, Y.~Rizk, and M.~Awad.
\newblock Damage identification in social media posts using multimodal deep
  learning.
\newblock In \emph{ISCRAM}. Rochester, NY, USA, 2018.

\bibitem[Netzer(2011)]{svhn}
e.~a. Netzer, Yuval.
\newblock {Reading digits in natural images with unsupervised feature
  learning.}
\newblock In \emph{Advances in Neural Information Processing Systems 32}, 2011.

\bibitem[Nguyen(2022)]{c:7}
e.~a. Nguyen, John.
\newblock {Federated learning with buffered asynchronous aggregation.}
\newblock In \emph{International Conference on Artificial Intelligence and
  Statistics. PMLR}, 2022.

\bibitem[Reddi et~al.(2020)Reddi, Charles, Zaheer, Garrett, Rush,
  Kone{\v{c}}n{\`y}, Kumar, and McMahan]{reddi2020adaptive}
S.~Reddi, Z.~Charles, M.~Zaheer, Z.~Garrett, K.~Rush, J.~Kone{\v{c}}n{\`y},
  S.~Kumar, and H.~B. McMahan.
\newblock Adaptive federated optimization.
\newblock \emph{arXiv preprint arXiv:2003.00295}, 2020.

\bibitem[Reiss and Stricker(2012)]{reiss2012introducing}
A.~Reiss and D.~Stricker.
\newblock Introducing a new benchmarked dataset for activity monitoring.
\newblock In \emph{2012 16th international symposium on wearable computers},
  pages 108--109. IEEE, 2012.

\bibitem[Ro et~al.(2022)Ro, Breiner, McConnaughey, Chen, Suresh, Kumar, and
  Mathews]{ro2022scaling}
J.~H. Ro, T.~Breiner, L.~McConnaughey, M.~Chen, A.~T. Suresh, S.~Kumar, and
  R.~Mathews.
\newblock Scaling language model size in cross-device federated learning.
\newblock \emph{arXiv preprint arXiv:2204.09715}, 2022.

\bibitem[Shi et~al.(2020)Shi, Li, Wang, Chen, Ye, and Xu]{shi2020hysync}
G.~Shi, L.~Li, J.~Wang, W.~Chen, K.~Ye, and C.~Xu.
\newblock Hysync: Hybrid federated learning with effective synchronization.
\newblock In \emph{2020 IEEE 22nd International Conference on High Performance
  Computing and Communications; IEEE 18th International Conference on Smart
  City; IEEE 6th International Conference on Data Science and Systems
  (HPCC/SmartCity/DSS)}, pages 628--633. IEEE, 2020.

\bibitem[Tian et~al.(2021)Tian, Chen, Yu, and Liao]{tian2021towards}
P.~Tian, Z.~Chen, W.~Yu, and W.~Liao.
\newblock Towards asynchronous federated learning based threat detection: A
  dc-adam approach.
\newblock \emph{Computers \& Security}, 108:\penalty0 102344, 2021.

\bibitem[Vaizman et~al.(2017)Vaizman, Ellis, and
  Lanckriet]{vaizman2017recognizing}
Y.~Vaizman, K.~Ellis, and G.~Lanckriet.
\newblock Recognizing detailed human context in the wild from smartphones and
  smartwatches.
\newblock \emph{IEEE pervasive computing}, 16\penalty0 (4):\penalty0 62--74,
  2017.

\bibitem[Wang(2022)]{c:26}
e.~a. Wang, Qiyuan.
\newblock {AsyncFedED: Asynchronous Federated Learning with Euclidean Distance
  based Adaptive Weight Aggregation}.
\newblock In \emph{arXiv preprint}, 2022.

\bibitem[Wang et~al.(2021)Wang, Charles, Xu, Joshi, McMahan, Al-Shedivat,
  Andrew, Avestimehr, Daly, Data, et~al.]{wang2021field}
J.~Wang, Z.~Charles, Z.~Xu, G.~Joshi, H.~B. McMahan, M.~Al-Shedivat, G.~Andrew,
  S.~Avestimehr, K.~Daly, D.~Data, et~al.
\newblock A field guide to federated optimization.
\newblock \emph{arXiv preprint arXiv:2107.06917}, 2021.

\bibitem[Wang et~al.(2020)Wang, Zhu, Chang, Shen, and Ren]{wang2020model}
Y.~Wang, T.~Zhu, W.~Chang, S.~Shen, and W.~Ren.
\newblock Model poisoning defense on federated learning: A validation based
  approach.
\newblock In \emph{International Conference on Network and System Security},
  pages 207--223. Springer, 2020.

\bibitem[Wu et~al.(2023)Wu, Zhang, Yu, Liu, Gu, Zhou, Chen, and
  Cheng]{wu2023personalized}
Y.~Wu, S.~Zhang, W.~Yu, Y.~Liu, Q.~Gu, D.~Zhou, H.~Chen, and W.~Cheng.
\newblock Personalized federated learning under mixture of distributions.
\newblock In \emph{International Conference on Machine Learning}, pages
  37860--37879. PMLR, 2023.

\bibitem[Xiao et~al.(2017)Xiao, Rasul, and Vollgraf]{xiao2017fashion}
H.~Xiao, K.~Rasul, and R.~Vollgraf.
\newblock Fashion-mnist: a novel image dataset for benchmarking machine
  learning algorithms.
\newblock \emph{arXiv preprint arXiv:1708.07747}, 2017.

\bibitem[Xie and Gupta.(2019)]{c:2}
S.~K. Xie, Cong and I.~Gupta.
\newblock {Asynchronous federated optimization}.
\newblock In \emph{arXiv preprint}, 2019.

\bibitem[Yang(2019)]{c:19}
e.~a. Yang, Ziqi.
\newblock {Neural network inversion in adversarial setting via background
  knowledge alignment.}
\newblock In \emph{Proceedings of the 2019 ACM SIGSAC Conference on Computer
  and Communications Security.}, 2019.

\bibitem[Ye(2020)]{c:20}
e.~a. Ye, Jingwen.
\newblock {Data-free knowledge amalgamation via group-stack dual-gan.}
\newblock In \emph{Proceedings of the IEEE/CVF Conference on Computer Vision
  and Pattern Recognition}, 2020.

\bibitem[Yin(2020)]{c:18}
e.~a. Yin, Hongxu.
\newblock {Dreaming to distill: Data-free knowledge transfer via
  deepinversion.}
\newblock In \emph{Proceedings of the IEEE/CVF Conference on Computer Vision
  and Pattern Recognition}, 2020.

\bibitem[Yin(2021)]{c:10}
e.~a. Yin, Hongxu.
\newblock {See through gradients: Image batch recovery via gradinversion.}
\newblock In \emph{Proceedings of the IEEE/CVF Conference on Computer Vision
  and Pattern Recognition}, 2021.

\bibitem[YYin(2020)]{c:21}
e.~a. YYin, Hongxu.
\newblock {Dreaming to distill: Data-free knowledge transfer via
  deepinversion.}
\newblock In \emph{Proceedings of the IEEE/CVF Conference on Computer Vision
  and Pattern Recognition.}, 2020.

\bibitem[Zhang(2018)]{c:23}
e.~a. Zhang, Richard.
\newblock {The unreasonable effectiveness of deep features as a perceptual
  metric.}
\newblock In \emph{Proceedings of the IEEE conference on computer vision and
  pattern recognition}, 2018.

\bibitem[Zhao(2018)]{c:1}
e.~a. Zhao, Yue.
\newblock {Federated learning with non-iid data}.
\newblock In \emph{arXiv preprint}, 2018.

\bibitem[Zhao and Bilen.(2020)]{c:12}
K.~R.~M. Zhao, Bo and H.~Bilen.
\newblock {idlg: Improved deep leakage from gradients.}
\newblock In \emph{arXiv preprint arXiv:2001.02610}, 2020.

\bibitem[Zheng et~al.(2017)Zheng, Meng, Wang, Chen, Yu, Ma, and
  Liu]{zheng2017asynchronous}
S.~Zheng, Q.~Meng, T.~Wang, W.~Chen, N.~Yu, Z.-M. Ma, and T.-Y. Liu.
\newblock Asynchronous stochastic gradient descent with delay compensation.
\newblock In \emph{International Conference on Machine Learning}, pages
  4120--4129. PMLR, 2017.

\bibitem[Zhou(2021)]{c:3}
e.~a. Zhou, Chendi.
\newblock {TEA-fed: time-efficient asynchronous federated learning for edge
  computing.}
\newblock In \emph{Proceedings of the 18th ACM International Conference on
  Computing Frontiers}, 2021.

\bibitem[Zhou and Lv.(2021)]{c:4}
Q.~Y. Zhou, Yuhao and J.~Lv.
\newblock {Communication-efficient federated learning with compensated
  overlap-fedavg.}
\newblock In \emph{IEEE Transactions on Parallel and Distributed Systems},
  2021.

\bibitem[Zhou et~al.(2021)Zhou, Ye, and Lv]{zhou2021communication}
Y.~Zhou, Q.~Ye, and J.~Lv.
\newblock Communication-efficient federated learning with compensated
  overlap-fedavg.
\newblock \emph{IEEE Transactions on Parallel and Distributed Systems},
  33\penalty0 (1):\penalty0 192--205, 2021.

\bibitem[Zhu et~al.(2022)Zhu, Kuang, Yang, and Qian]{zhu2022client}
H.~Zhu, J.~Kuang, M.~Yang, and H.~Qian.
\newblock Client selection with staleness compensation in asynchronous
  federated learning.
\newblock \emph{IEEE Transactions on Vehicular Technology}, 72\penalty0
  (3):\penalty0 4124--4129, 2022.

\bibitem[Zhu et~al.(2019)Zhu, Liu, and Han]{zhu2019deep}
L.~Zhu, Z.~Liu, and S.~Han.
\newblock Deep leakage from gradients.
\newblock \emph{Advances in neural information processing systems}, 32, 2019.

\bibitem[Zhu et~al.(2021)Zhu, Hong, and Zhou]{zhu2021data}
Z.~Zhu, J.~Hong, and J.~Zhou.
\newblock Data-free knowledge distillation for heterogeneous federated
  learning.
\newblock In \emph{International conference on machine learning}, pages
  12878--12889. PMLR, 2021.

\bibitem[Zhu and Han.(2019{\natexlab{a}})]{c:5}
Z.~L. Zhu, Ligeng and S.~Han.
\newblock {Deep leakage from gradients.}
\newblock In \emph{Advances in neural information processing systems},
  2019{\natexlab{a}}.

\bibitem[Zhu and Han.(2019{\natexlab{b}})]{c:9}
Z.~L. Zhu, Ligeng and S.~Han.
\newblock {Deep leakage from gradients}.
\newblock In \emph{Advances in neural information processing systems 32},
  2019{\natexlab{b}}.

\end{thebibliography}

\newpage
\appendix

\section{A: Evaluations with other data modalities}
In the experimental section of the main text, we conduct experiments on three benchmark CV datasets and one real-world CV dataset. To demonstrate the generality of our method, we further conduct experiments on two human activity recognition (HAR) datasets with time-series data:
\begin{itemize}
	\item \textbf{PAMAP2 \cite{reiss2012introducing}} with 13 classes of human activities and over 2M data samples collected using IMU and heart rate sensors. A 3-layer MLP model is used in FL.
	\item \textbf{ExtraSensory \cite{vaizman2017recognizing}} with over 300k data samples collected using IMU, gyroscope and magnetometer sensors on smartphones. Besides 7 main labels of activities (e.g., standing, laying down, etc), it also provides 109 additional labels describing more specific activity contexts. An 1D-CNN model is used in FL.
\end{itemize}  
We set three levels of staleness to 2, 5, and 10 epochs, while other settings remain the same as those in the main results. As shown in Table \ref{appendix:har}, our approach demonstrates an advantage under medium or high staleness.

\begin{table*}[ht]
	\centering
	%  %\vspace{-0.05in}
	\begin{tabular}{|c||c|c|c||c|c|c||}
		\hline
		&\multicolumn{3}{c||}{\textbf{PAMAP2 / MLP}} & \multicolumn{3}{c||}{\textbf{ExtraSensory / 1D-CNN}}  \\
		\hline
		staleness	& low & medium & high  & low & medium & high\\
		\hline \hline
		Unweighted & 0.0\%& 0.0\%&0.0\%           & 0.0\% &0.0\% &0.0\%             \\
		\hline
		Weighted   & -5.4\% &-13.9\% &-43.5\% &-18.4\% &-46.5\% &-62.3\% \\
		\hline
		Asyn-tiers &+0.7\% & +0.4\%& -0.5\%  &-2.0\% &+0.8\% & -2.9\%   \\
		\hline
		1st-Order  &+2.3\% &+1.5\%&+0.6\%   &+3.6\% &+2.5\% &-2.2\%    \\
		\hline
		W-Pred     &+2.6\% &+1.3\% &+0.6\%   &+0.4\% & +1.5\%&-1.3\%    \\
		\hline\hline
		\textbf{Ours}       &-1.9\% &+5.4\%  &70.6\% &-3.0\% & +16.9\%&+34.2\%    \\
		\hline
		
	\end{tabular}
	%\vspace{0.05in}
	\caption{The trained model's accuracy in data classes affected by device delays, with different amounts of device delays. Accuracy is shown as the relative improvement compared to unweighted aggregation.}  
	\label{appendix:har}
	%\vspace{-0.1in}
\end{table*}

Besides, our method can also be applied to tasks involving other data modalities, such as text. Since in NLP, text is decomposed into discrete tokens, we must estimate data in the continuous embedding space \cite{zhu2019deep}. Since errors occur when projecting the estimated data from the embedding space into discrete tokens, text is more prone to gradient inversion attacks, requiring prior knowledge for a successful attack \cite{gupta2022recovering,dimitrov2022lamp}. This suggests that when applied to test data, the privacy leakage risk of our method would be lower.

\section{B: Comprehensive evaluation on weighted aggregation}
Applying a smaller weight to a stale update can reduce the error introduced in Federated training. However, under intertwined heterogeneities, applying reduced weights to stale updates degrades the accuracy of data samples affected by these heterogeneities. This occurs because the contributions of these data to the global model are also reduced, so the trained global model contains less knowledge about these data.

Intuitively, if we increase the weight of stale updates, their contributions are forced to increase, leading to higher accuracy on the stale clients. However, the errors in the stale updates are also magnified and incorporated into the global model, which decreases the model's accuracy on other data. To verify this, we simulate an FL system with 100 clients, 10 of which are stale, and train a LeNet model on the MNIST dataset. As shown in Table \ref{acc_weight}, compared to unweighted updates, applying increased weight improves the accuracy on the 10 stale clients by around 10\%, but decreases the overall accuracy across all 100 clients by around 5\%. Clearly, such a trade-off is unacceptable. Therefore, we should compensate the error in the stale updates instead of exploring weighting strategies.

\begin{table}[h]
        \small
	\centering
%	%\vspace{-0.05in}
	\begin{tabular}{c||ccc}
 
		\hline
		Weighting strategy  & Reduced W &Non & Increased W\\
		\hline 
            \hline  
            Acc - stale clients (\%)       &39.2 &57.4 & 68.1  \\
            \hline  
		Acc - all clients (\%)  &81.4 &80.5 &75.4\\
		\hline
	\end{tabular}
        %\vspace{-0.05in}
	\caption{Model accuracy under different weighting strategies}	
	\label{acc_weight}
	%\vspace{-0.1in}
\end{table}

\begin{table*}[ht]
	\centering
	\begin{tabular}{c|c|c|c}
		\hline
		Method & GI based estimation & direct  aggregation & using samples from generative model   \\ \hline\hline
		Estimation error  & 0.32  & 0.52  & 0.86   \\ \hline
	\end{tabular}	
	\caption{Error of estimating the non-stale model updates with different data recovery methods, measured by 1-cosine similarity}
	\label{tab:error}
\end{table*}

\section{C: Other approaches to estimating the data knowledge}
Our basic approach in this paper is to use the gradient inversion technique \cite{c:5,c:12} to estimate knowledge about the clients' local training data from their uploaded stale model updates, and then use such estimated knowledge to compute the corresponding non-stale model updates for aggregation in FL. In this section, we provide supplementary justifications about the ineffectiveness of other methods for such data knowledge estimation, hence better motivating our proposed design.
\begin{figure}[ht]
	\centering
	\includegraphics[width=0.8\columnwidth]{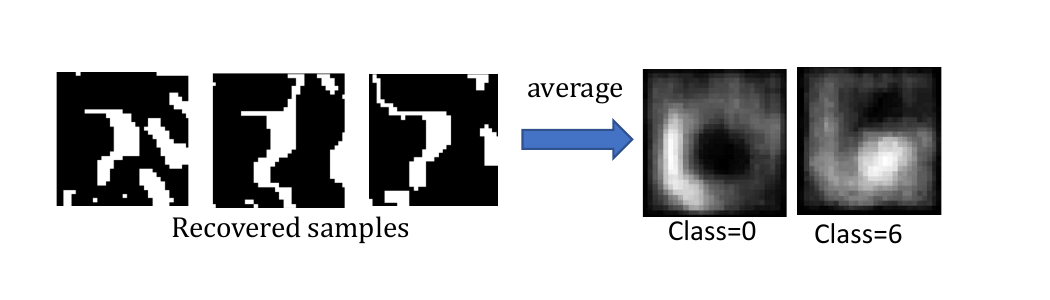} % 
	\caption{The computed data samples, when being averaged, can represent a blurred image that matches the data pattern in the original dataset (i.e., handwriting digits in the MNIST dataset)}
	\label{average_rec}
\end{figure}

The most commonly used approach to recovering the training data from a trained ML model involves training an extra generative model, to compel its generated data samples to exhibit high predictive value on the original model \cite{zhu2021data}, and to add image prior constraint terms to enhance data quality \cite{c:22}. On the other hand, data recovery can also be achieved by directly optimizing randomly initialized input data until it performs well on the original model \cite{c:21}. However, the results of our preliminary experiments, using the LeNet model and and the MNIST dataset, show that none of these approaches can provide good quality of the computed data, in order to be used in our FL scenarios.

More specifically, these existing approaches can ensure that the computed dataset, as a whole, exhibits some characteristics of the original training data. For example, as shown in Figure \ref{average_rec}, the averaged image of the recovered data samples in each data class can resemble a meaningful image that matches the data pattern in the original dataset. However, the individual image samples being computed have very low quality. If these computed data samples are used to compute the non-stale model updates in FL, it will result in a significant error in estimating the non-stale model updates, which greatly exceeds the error produced by our proposed gradient inversion (GI) based estimation, as shown in Table \ref{tab:error}.

\begin{figure}[ht]
	\centering
	\includegraphics[width=0.6\columnwidth]{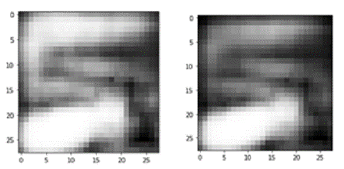} % 
	\caption{Two computed image samples in class 5 of the MNIST dataset}
	%\vspace{-0.1in}
	\label{similar_image}
\end{figure}
Furthermore, as shown in Figure \ref{similar_image}, the computed data samples lack diversity, resulting in high similarity among the generated samples in the same data class. Training on such highly similar data samples can easily lead the model to overfit. 
\begin{figure}[ht]
	\centering
	\includegraphics[width=0.6\columnwidth]{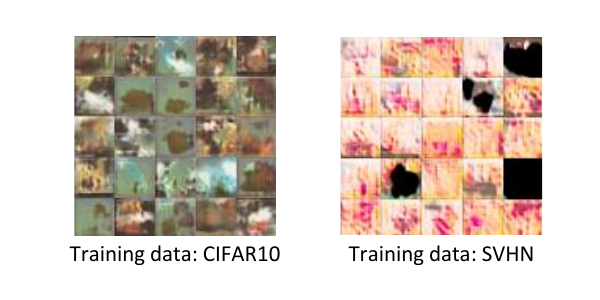} % 
	\caption{The computed data samples from different training datasets}
	\label{public}
\end{figure}
Some attempts have been made to enhance the data quality by incorporating another extra public dataset to introduce general image knowledge \cite{c:19}, but the effectiveness of this approach highly depends on the specific choice of such public dataset. Experimental results in \cite{jeon2021gradient} demonstrate that the quality of computed data can only be ensured if the extra public dataset shares the similar data pattern with the original training dataset. For example, in our preliminary experiments, we selected CIFAR-100 as the public dataset, and the original training datasets included CIFAR-10 and SVHN datasets. As shown in Figure \ref{public}, when CIFAR-10 is used as the training dataset, the computed data exhibits higher quality compared with that using the SVHN dataset as the training dataset, because the CIFAR-10 dataset shares the similar image patterns with the CIFAR-100 dataset.

Compared to the existing methods, our proposed technique uses gradient inversion to obtain an estimation of the clients' original training data. Since we only require the computed data to mimic the model's gradient produced with the original training data, we do not necessitate the quality of individual data samples being computed, and could hence avoid the impact of the computed data's low quality on the FL performance.
\begin{figure}[ht]
	\centering
	%\vspace{-0.1in}
	\includegraphics[width=0.6\columnwidth]{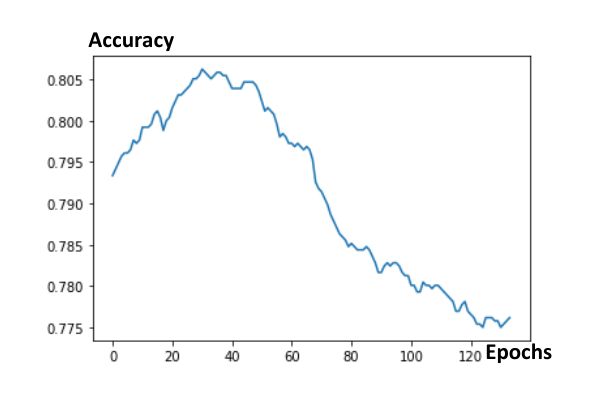} % 
	\caption{The knowledge distillation, when being applied to retrain the global model in FL, will overfit after a specific number of epochs}.
	%\vspace{-0.1in}
	\label{overfit}
\end{figure}

Because of such low quality of the computed data, they cannot be directly used to retrain the global model in FL, as an estimate to non-stale model updates. Some existing approaches, instead, use knowledge distillation to transfer the knowledge contained in the computed data to the target ML model \cite{zhu2021data}. However, in our FL scenario, since the server only conducts aggregation of the received clients' model updates and lacks the corresponding test data (as part of the clients' local data), the server will be unable to decide if and when the model retraining will overfit (Figure \ref{overfit}). Furthermore, all the existing methods require that the clients' model updates have to be fully trained, but this requirement generally cannot be satisfied in FL scenarios.

\section{D: Hyper-parameters in gradient inversion}
\textbf{Deciding the size of $D_{rec}$:} In our proposed approach to estimating the clients' local data distributions from stale model updates, a key issue is how to decide the proper size of $D_{rec}$. Since gradient inversion is equivalent to data resampling in the original training data's distribution, a sufficiently large size of $D_{rec}$ would be necessary to ensure unbiased data sampling and sufficient minimization of gradient loss through iterations. On the other hand, when the size of $D_{rec}$ is too large, the computational overhead of each iteration would be unnecessarily too high.

We experimentally investigated such tradeoff by using the MNIST and CIFAR-10 \cite{cifar10} datasets to train a LeNet model. Results in Tables \ref{tab:GI_loss_mnist} and \ref{tab:GI_loss_cifar}, where the size of $D_{rec}$ is represented by its ratio to the size of original training data, show that when the size of $D_{rec}$ is larger than 1/2 of the size of the original training data, further increasing the size of $D_{rec}$ only results in little extra reduction of the gradient inversion loss but dramatically increase the computational overhead. Hence, we believe that it is a suitable size of $D_{rec}$ for FL. Considering that  clients' local dataset in FL contain at least hundreds of samples, we expect a big size of $D_{rec}$ in most FL scenarios.

\begin{table}[H]
	\centering
	%%\vspace{-0.1in}
	\begin{tabular}{c||c|c|c|c|c|c}
		\hline
		Size & 1/64 & 1/16 & 1/4  & 1/2  & 2    & 10   \\ \hline\hline
		Time(s) & 193  & 207  & 214  & 219  & 564  & 2097 \\ \hline
		GI loss & 27   & 4.1  & 2.56 & 1.74 & 1.62 & 1.47 \\ \hline
	\end{tabular}
	%\vspace{0.1in}
	\caption{Tradeoff between gradient inversion (GI) loss and computing time with different sizes of $D_{rec}$ after 15k iterations, with the MNIST dataset}
	\label{tab:GI_loss_mnist}
	%\vspace{-0.1in}
\end{table}

\begin{table}[H]
%	%\vspace{-0.1in}
	\centering
	\begin{tabular}{c||c|c|c|c|c|c}
		\hline
		Size & 1/64 & 1/16 & 1/4  & 1/2  & 2    & 10   \\ \hline\hline
		Time(s) & 423  & 440  & 452  & 474  & 1330  & 4637 \\ \hline
		GI Loss & 1.97   & 0.29  & 0.16 & 0.15 & 0.15 & 0.12 \\ \hline
	\end{tabular}	
%\vspace{0.1in}
	\caption{Tradeoff between gradient inversion (GI) loss and computing time with different sizes of $D_{rec}$ after 15k iterations, with the CIFAR-10 dataset}
	\label{tab:GI_loss_cifar}
	%\vspace{-0.1in}
\end{table}

\textbf{Deciding the metric for model difference:} Such a big size of $D_{rec}$ directly decides our choice of how to evaluate the change of $w^{t-\tau}_i$ in Eq. (2). Most existing works use cosine similarity between $Local Update (w^{t-\tau}_{global};D_{rec})$ and $w^{t-\tau}_i$ to evaluate their difference in the direction of gradients, so as to maximize the quality of individual data samples in $D_{rec}$ \cite{c:28}. However, since we aim to compute a large $D_{rec}$, this metric is not applicable, and instead we use L1-norm as the metric to evaluate how using $D_{rec}$ to retrain $w^{t-\tau}_{global}$ will change its magnitude of gradient, to make sure that $D_{rec}$ incurs the minimum impact on the state of training. 

\section{E: Gradient inversion under diverse FL settings}
In the main text of the paper, we empirically verified that our gradient inversion-based compensation achieves significantly smaller error compared to first-order compensation in a simple FL setting. However, in FL, many factors can affect the training process. Therefore, in this section, we further evaluate the error under diverse settings to demonstrate that our methods have broad applications in real-world FL systems.

The first factor we consider is the number of steps in local training, as existing works \cite{c:11} indicate that a complex local training program makes gradient inversion more difficult. Under such a high degree of data heterogeneity, the divergence between the client model and the global model increases with the number of local steps\cite{c:17}, making compensation more challenging. We use the LeNet model, the MNIST dataset, and an SGD optimizer to compute a stale update and apply different methods to compensate for it. The results are shown in Table \ref{erorr_iter}. Although the error of our method increases with the number of steps, it remains much smaller than that of the first-order method.

\begin{table}[h]
	\centering
%	%\vspace{-0.05in}
	\begin{tabular}{c||cccccc}
		\hline
		\# of iterations  & 1 & 5 & 10 & 20 &50\\
		\hline 
            \hline  
            GI method       &0.05 &0.18 & 0.22 & 0.22 &0.26\\
            \hline  
		1st-order method &0.14 &0.31 &0.33& 0.35 &0.38\\
		\hline
	\end{tabular}
        %\vspace{-0.05in}
	\caption{Compensation error (measured using L1-norm distance) under different numbers of iterations in client's local training program}	
	\label{erorr_iter}
	%\vspace{-0.1in}
\end{table}

Except for basic SGD, various optimizers are used in different FL systems. We test our methods with four optimizers: SGD, SGD with momentum (SGDM), Adam, and FedProx, where FedProx is an optimization method designed for FL with data heterogeneity by using a proximal term \cite{li2020federated}. As shown in Table \ref{error_optimizer}, our method can achieve a smaller compensation error with most optimizers. Although our method fails when using adaptive optimizers like Adam, in our practice, under a high degree of data heterogeneity, it's not recommended to use these adaptive optimizers for training stability.

\begin{table}[h]
	\centering
%	%\vspace{-0.05in}
	\begin{tabular}{c||ccccc}
		\hline
		Optimizer  & SGD & SGDM & Adam & FedProx\\
		\hline 
            \hline  
            GI method       &0.22 &0.26 & 0.44 & 0.17 \\
            \hline  
		1st-order method &0.33 &0.35 &0.38& 0.30\\
		\hline
	\end{tabular}
        %\vspace{-0.05in}
	\caption{Compensation error (measured using L1-norm distance) under different optimizers in client's local training program}	
	\label{error_optimizer}
	%\vspace{-0.1in}
\end{table}

\section{F: Error caused by gradient sparsification}
In the main text of the paper, we showed that with 95\% sparsification, we can reduce computation and protect privacy with only a small increase in estimation error. To better evaluate the trade-off between performance and efficiency/privacy, we further compare the training accuracy at different rates of sparsification using LeNet and MNIST dataset. As shown in Table \ref{sparse_acc}, with a sparsification rate of 95\%, the accuracy drop compared to non-sparsification is minor, but is high enough to achieve computation savings and privacy protection. Further increase the sparsification rate will help reduce more computation in the gradient inversion, but the accuracy drop is significant.
\begin{table}[h]
	\centering
%	%\vspace{-0.05in}
	\begin{tabular}{c||ccccc}
		\hline
		Sparsification rate & 0\% & 90\% & 95\% & 99\%\\
		\hline 
            \hline  
            Accuracy       &63.5 &61.9 & 61.2 & 53.3 \\
            \hline  
	\end{tabular}
        %\vspace{-0.05in}
	\caption{Model accuracy with different rates of sparsification}	
	\label{sparse_acc}
	%\vspace{-0.1in}
\end{table}

\section{G: Other Discussions}

\subsection{Server's overhead in large-scale FL systems}
In large-scale FL systems, the server must compute gradients for each delayed client, potentially leading to a performance bottleneck. However, our method requires applying gradient inversion only to a subset of stale model updates, which encapsulate unique and critical knowledge absent from other unstable updates. We argue that, in most practical scenarios as we listed in the main text of the paper, the volume of these updates is likely to remain small, even in large-scale FL systems.

On the other hand, most of current FL implementations, such as \cite{charles2021large}, usually use a variant client selection rate, so that the number of clients participating in one global round remains constant instead of increasing proportionally with the total number of clients. Essentially, existing work \cite{ro2022scaling} showed that once the number of clients per global round is sufficiently large (e.g., 10-50 even for FL systems with thousands of clients), further increasing such number yields only marginal performance gains but significantly increases overhead, and could also result in catastrophic training failure and generalization failure. Hence, our approach will not significantly increase the server’s computing overhead in large-scale FL systems.

\subsection{Defense against malicious attackers}
In practical FL systems, a malicious attacker may intentionally inject abrupt gradients (e.g., with extremely large or small values) into the server. Such attacks not only disrupt gradient inversion but also undermine the FL training process itself, hindering convergence and reducing model accuracy. While defending against such attacks is not the primary focus of this paper, existing works have proposed defenses against these gradient-based attacks \cite{wang2020model}.

\subsection{Applying statistical privacy methods}
Since our method only modifies the FL operations on the server and keeps other FL steps (e.g., the clients' local model updates and client-server communication) unchanged, local differential privacy can theoretically be directly applied to our approach without any modification. More specifically, each client can independently add Gaussian noise to its local model updates, before sending the updates to the server \cite{geyer2017differentially}. 

Moreover, note that differential privacy (DP) is also orthogonal to our proposed privacy protection method, because noise can be added to the gradient after our proposed sparsification method.

\end{document}